\documentclass[10pt,journal,compsoc]{IEEEtran}

\usepackage{times}
\usepackage{epsfig}
\usepackage{graphicx}
\usepackage{amsmath}
\usepackage{amssymb}
\usepackage[ruled,vlined]{algorithm2e}
\usepackage{algpseudocode}
\usepackage{caption}
\usepackage{subfigure}
\usepackage{booktabs}
\usepackage{tabularx}
\usepackage{xcolor}

\usepackage{cite}
\usepackage{textcomp}
\usepackage{mathrsfs}
\usepackage{graphicx}
\graphicspath{ {./images/} }

\begin{document}
	%
	\title{AirFi: Empowering WiFi-based Passive Human Gesture Recognition to Unseen Environment via Domain Generalization}
	%
	%
	%
	
	\author{Dazhuo Wang,
		Jianfei Yang$^*$,
		Wei Cui,
		Lihua~Xie,~\IEEEmembership{Fellow,~IEEE},
		and Sumei Sun,~\IEEEmembership{Fellow,~IEEE}
		\thanks{D. Wang, J. Yang and L. Xie are with the School of Electrical and Electronics Engineering, Nanyang Technological University, Singapore (E-mail: dazhuo001@e.ntu.edu.sg; yang0478@ntu.edu.sg; elhxie@ntu.edu.sg).}
		\thanks{W. Cui and S. Sun are with the Institute for Infocomm Research (I$^2$R), Agency for
			Science, Technology and Research (A*STAR), Singapore 138632.(E-mail: Cui\_Wei@i2r.a-star.edu.sg; sunsm@i2r.a-star.edu.sg).}
		\thanks{$^*$J. Yang is the corresponding author (yang0478@ntu.edu.sg).}
	}

	\markboth{IEEE Transactions on Mobile Computing}%
	{Shell \MakeLowercase{\textit{et al.}}: Bare Demo of IEEEtran.cls for IEEE Journals}

	\maketitle
	
	\begin{abstract}

WiFi-based smart human sensing technology enabled by Channel State Information (CSI) has received great attention in recent years. However, CSI-based sensing systems suffer from performance degradation when deployed in different environments. Existing works solve this problem by domain adaptation using massive unlabeled high-quality data from the new environment, which is usually unavailable in practice. In this paper, we propose a novel augmented environment-invariant robust WiFi gesture recognition system named AirFi that deals with the issue of environment dependency from a new perspective. The AirFi is a novel domain generalization framework that learns the critical part of CSI regardless of different environments and generalizes the model to unseen scenarios, which does not require collecting any data for adaptation to the new environment. AirFi extracts the common features from several training environment settings and minimizes the distribution differences among them. The feature is further augmented to be more robust to environments. Moreover, the system can be further improved by few-shot learning techniques. Compared to state-of-the-art methods, AirFi is able to work in different environment settings without acquiring any CSI data from the new environment. The experimental results demonstrate that our system remains robust in the new environment and outperforms the compared systems.
	\end{abstract}
	
	
	
	\section{Introduction}

With the increasing popularity of WiFi technology and wide availability of WIFi infrastructure, WiFi-based smart human sensing is receiving more attention and plays an important role in Internet of Things (IoT) systems. It has many applications in health care, security, entertainment and tailored services \cite{8941292,9115830,8519328,7345587}.

WiFi-based sensing technology is enabled by Channel State Information (CSI) \cite{yang2022deep,halperin2010predictable,hu2017new,halperin2011tool,zou2017freecount} which is a fine-grained measurement at the physical layer from a subcarrier channel. It can reflect the complex multipath effect caused by human motion due to its frequency diversity and capture multipath propagation of WiFi signal from transmitters to receivers over multiple subcarriers, which reveals various influences of human motion. As a result, different human behaviors can be detected and recognized by observing the variations of CSI data~\cite{zou2017freedetector,zou2017multiple,zou2017poster,zou2019wifi,zou2018wifi}. Compared with other existing smart human sensing technologies using either vision-based sensing techniques or wearable sensor based techniques, WiFi-based smart sensing has many advantages. It does not require either line of sight or illumination. Besides, it leverages the existing WiFi infrastructure in buildings and at homes, which reduces the additional cost of wearable devices and is more user-friendly. It also provides better privacy protection than camera based solutions. 

Though existing CSI-based smart human sensing systems have achieved decent performances in some applications, many of them still suffer from serious performance degradation under multiple environments~\cite{zou2018robust,zou2018joint}. A high-accuracy system trained in one environment cannot be readily deployed in another environment due to the performance degradation caused  by the different environment setting.

There are currently two main methods to address this problem: model based method and learning based method. Model based method aims to build a signal model using extracted signal parameters that suffer the least perturbation under changing environment dynamics \cite{zhang2017toward,wang2016human,wang2015understanding,pu2013whole}. However, it requires expert knowledge for parameter selection. Instead of using deep learning networks to learn the data features directly from the CSI data, model based methods manually select signal parameters from received signal data to construct system models. However, to decide which parameters to choose requires expert knowledge in signal transmission fields. It increases the difficulty of CSI based human behavior recognition system design. Meanwhile, by selecting only parts of the signal information from the complete CSI sequences, it may cause loss of some important features. Most importantly, it is hard for model based methods to recognize complex human behaviors, such as gestures due to their weak fitting ability. Recently learning based method has drawn increasing attention with the development of deep learning techniques. One popular technique applied in the CSI-based human behavior recognition is Domain Adaption (DA) \cite{daume2009frustratingly,wang2018deep,pan2010domain}. DA is able to transfer a CSI-based sensing system trained in one environment (source domain) to another environment (target domain)~\cite{zou2019consensus,yang2020mind,yang2020mobileda}. But the DA based CSI human behavior sensing systems require a large number of CSI samples from the new environment to perform domain adaption, which is not practical in many scenarios. In \cite{wang2021multimodal}, simulated fake data is used to solve this issue, however, an adequate amount of CSI data from the new environment are still needed to generate fake data. In order to address this problem, we aim to use the CSI data from multiple training environments to train a generalized system so that the model can be applied to a new testing environment without collecting any CSI data from the testing environment.

In this paper, we propose a novel Augmented environment-Invariant Robust WiFi gesture recognition system AirFi that aims to solve the performance degradation of a CSI-based smart sensing system in unseen environments. In our scenario, CSI data from the testing environment is not available. AirFi addresses this issue by generalizing its system model to the testing environment. We are inspired by the idea of Domain Generalization (DG) \cite{muandet2013domain,li2018learning,zhou2021domain}. AirFi is trained using CSI data from multiple training environmental settings in such a way that the model can generalize well to a new environment setting. Firstly, AirFi uses an encoder to extract feature codes from CSI data collected in several training environment settings. Then with the extracted features mapped on the feature space, AirFi minimizes the distribution differences between feature codes from different environment settings. Finally, the feature codes are used to train the classifier. In this way, AirFi is able to generalize its model to unseen environment settings. Besides, in order to enhance the system model training, an additional random prior distribution is introduced to the feature extraction process in an adversarial manner. It reduces the dependency between the model and training CSI data. Data augmentation and feature augmentation techniques are also applied to improve the system training. Experiments show that AirFi achieves decent performance and outperforms the benchmarking reference systems.

The contributions of the paper are summarized as follows:
\begin{itemize}
	
	\item We propose a CSI-based gesture recognition system AirFi that can generalize to a new environment without any new data by domain generalization. To the best of our knowledge, the AirFi is the first work that deals with the environment dependency issue without collecting new data or adapting the model in the new environment.
	\item For better generalization, we augment the CSI data and feature codes to improve their representativity. Unlike previous works which augment CSI data and features randomly, in this paper the augmentation is designed to be more aggressive on the domain direction while less aggressive on the class-wise direction using a label dependent regularizer.
	\item In the new environment, by applying the few-shot learning technique the performance of our proposed system AirFi can be further improved with a few CSI data from the testing environment setting. 
	\item Experiments show that our proposed system gives decent performances across different environments. With few-shot learning techniques, the performances are further improved.
\end{itemize}

The rest of the paper is organized as follows: Section II discusses related works. Section III provides the detail of AirFi system design. Section IV shows experimental results and comparisons with existing works. Section V concludes the paper and provides recommendations for some future research topics.

\section{RELATED WORKS}

In this section, we are going to review some previous works on CSI-based behavior recognition and their methods to overcome the environmental dynamics. Besides, we will also review some related works on DA and DG. Finally, some few-shot learning works will be reviewed.

\subsection{CSI based Human Behavior Recognition}
Human behavior recognition has been receiving great attention in recent years~\cite{li2021two,zhang2021privacy,9305332,yang2018fine,yang2018carefi,deng2022gaitfi,yang2022autofi,yang2022securesense,yang2022efficientfi}. Using large number of CSI samples, an accurate human behavior recognition system can be built. Reference \cite{xiong2015csi} proposes a system that uses the CSI to recognize human gestures. In \cite{tan2016wifinger}, the system model is further improved with an environmental noise removal mechanism to mitigate the effect of signal dynamics due to environment changes. Reference \cite{8514811} proposes a deep learning-based approach which is called attention based bi-directional long short-term memory (ABLSTM), for passive human activity recognition using WiFi CSI signals. WiGrus leverages the CSI to recognize a set of hand gestures using software defined radio \cite{8832182}. In \cite{yang2019learning}, the authors propose a novel deep Siamese representation learning architecture for one-shot gesture recognition. 

While the issue of environmental dynamics is becoming one of the most critical challenges faced by existing CSI-based human behavior sensing systems, many research works have studied this problem. Existing methods can be categorized into two main categories, model based methods and learning based methods.

Model based method manages to use signal parameters to build a signal model which suffers the least perturbation under changing environment dynamics. References \cite{zhang2017toward} and \cite{wang2016human} use Fresnel Zone for human respiration and walking detection. Carm \cite{wang2015understanding} uses both CSI-speed model and CSI-activity model to quantify the correlation between the movement speeds of different human body parts and specific human activities. Doppler shifts are measured and used for determining the directions of human motions in WiSee \cite{pu2013whole}.  A movement towards the receiver causes a positive Doppler shift, while a movement away from the receiver results in a negative shift. WiAnti, a CSI-based activity recognition system that addresses the issue of co-channel interference by using adaptive subcarrier selection is proposed in \cite{8726822}. Reference \cite{li2020location} proposes a location-free activity recognition system using Angle Difference of Arrival (ADoA). Widar3.0 \cite{zheng2019zero} combines both physical model and DNN learning to achieve the cross-domain gesture recognition. It builds the body coordinate profile for gesture recognitions. It requires the location and orientation of the person to calculate the domain-independent BVP which is not usually available in the real world situation. The system WiFH proposes \cite{9141400} a very interesting idea that one user tends to perform one specific gesture in the same way across different environments. It mentions that the performance of gesture recognition is improved when the user identity tasks and gesture recognition tasks are treated as a combined tasks. It uses the information of users’ identification to help the training process of gesture recognitions. Though model based methods achieve good performances in these aforementioned works, they have some limitations. First, expert knowledge is required for parameter selection. Besides, selecting only parts of the signal information may cause some important features missing. Lastly, it is hard for model based methods to recognize different kinds of human activities due to their weak fitting ability.

Learning based methods have received great attention in recent years \cite{yang2019learning,8514811,9305332,zou2018towards}. With the development of computer vision and deep learning algorithms, some works on CSI-based human sensing are highly inspired by some novel deep learning research topics. Some recent works apply the idea of DA from deep learning research. Reference \cite{9145101} proposes a novel scheme for CSI-based behavior recognition tasks that uses an activity filter-based deep learning network with enhanced correlation features to achieve robustness under different environmental settings. In \cite{9322627}, an environment-robust CSI-based human behavior sensing system is proposed. It leverages the properties of a matching network and enhanced features to create an environment-robust behavior recognition. Reference \cite{xiao2019csigan} adapts the idea of the generative adversarial network to perform the domain adaption for model transfer. Reference \cite{zhang2018crosssense} applies a roaming model which is also able to transfer the system model to a new environment using labeled data from the target environment. EI \cite{jiang2018towards} also utilizes data from both source and target domains to train the system model. It achieves good results with limited amount of unlabelled CSI data from the target domain.  DA transfers the system model trained in one environment (source domain) to be applied in a new different environment (target domain). DA, however, requires a large number of CSI samples from the new environment, which is not practical in many scenarios. MCBAR \cite{wang2021multimodal} and CSIGAN \cite{xiao2019csigan} use simulated fake data to mitigate this issue, however, an adequate amount of CSI data from the new environment is still needed to perform the domain transfer of the trained model. In this paper, we study the problem of human activity recognition with limited or no data from the target domain  with the idea of using CSI data from multiple training environments to train the system so that the system model can be better generalized to a new testing environment.

\subsection{Domain Adaption and Generalization}

In the last few years, great success has been achieved by machine learning. The corresponding works have benefited many real-world applications including the CSI-based human sensing field \cite{wang2018deep}. However, it takes a lot of resources to collect and annotate each dataset for new tasks. Especially when the number of samples and domains are very large, it can be an extremely resource-consuming and time-consuming process. Besides, sufficient data samples will not always be available in certain circumstances. For example, it is not user-friendly to collect large numbers of data from users when the systems are deployed in users' places. This motivates the research works on reusing a trained model in a new domain. DA is one of the methods proposed to achieve this goal.

Recent works focus on transferring network representations from the source domain where labeled data datasets are easy to acquire to a target domain where labeled data is sparse or even non-existent \cite{tzeng2017adversarial}. Reference \cite{tzeng2014deep} proposes a new CNN architecture that introduces an adaptation layer and an additional domain confusion loss, to learn a representation that is both semantically meaningful and domain invariant. In \cite{long2015learning}, a new Deep Adaptation Network architecture is proposed which generalizes deep convolutional neural network to the domain adaptation scenario. 
Reference \cite{liu2017unsupervised} makes a shared-latent space assumption and proposes an unsupervised image-to-image translation framework based on Coupled GANs. In \cite{huang2018multimodal}, a multimodal unsupervised image-to-image translation framework is proposed by assuming that the image representation can be decomposed into a content code that is domain-invariant, and a style code that captures domain-specific properties. Both \cite{liu2017unsupervised} and \cite{huang2018multimodal} can provide the ability that data from one domain can be transferred into another domain without changing the categories of the data samples. In \cite{liu2016coupled}, the CoGAN learns a joint distribution of images in the two domains from images drawn separately from the marginal distributions of the individual domains by enforcing a simple weight-sharing constraint. The main strategy is to guide feature learning by minimizing the difference between the source and target feature distributions. Some other methods also manage to minimize the Maximum Mean Discrepancy (MMD) loss for this purpose \cite{li2018domain}. DA has achieved great success and benefited many systems and applications. However, the limitation of DA is that it still needs many data samples from the target domain in order to perform the domain transfer of system models. In many other scenarios, there may not be any data of the target domain during the training phase, but the system is still needed to build a precise model for a totally new target domain. To address this problem, domain generalization is proposed.

DG leverages the labeled data from multiple source domains to learn a universal representation, which is expected to generalize well for an unseen target domain \cite{zhou2021domain}. DG is firstly introduced in \cite{blanchard2011generalizing}. They identify an appropriate reproducing kernel Hilbert space and optimize a regularized empirical risk over the space. Then in \cite{khosla2012undoing}, a discriminative framework is used to directly exploit dataset bias during training. In \cite{goodfellow2014generative}, they propose a new framework for estimating generative models via an adversarial process using a generative model and a discriminative model. Reference \cite{li2015generative} utilizes MMD, which leads to a simple objective that can be interpreted as matching all orders of statistics between datasets and samples from the model. Reference \cite{li2018deep} leverages deep neural networks for domain-invariant representation learning and achieves end-to-end conditional invariant deep domain generalization.  The above works inspire us to use several environments where data is relatively easier to collect as source domains. Then we generalize and apply the trained model into a new environment which is regarded as the target domain. As it is difficult to collect large numbers of CSI data to train a generalized system model in our research problem, we utilize the data and feature augmentation techniques to improve the model generalization.

\subsection{Few-shot Learning}

Machine learning has been highly successful in data-intensive applications, but is often hampered when the data set is small \cite{wang2020generalizing}. It reduces its scalability to new classes. To address this problem, few-shot learning is proposed to tackle this problem. Few-shot learning aims to train a system model using very few labeled samples.

Rather than traditional convolutional neural networks which need a large number of data to train the network layers, few-shot learning techniques manage to build a model with only a limited amount of data. 

In references \cite{goldberger2004neighbourhood,salakhutdinov2007learning,weinberger2009distance,min2009deep}, neighborhood component analysis is applied. They optimize the K Nearest Neighbor accuracy in the feature space with limited amounts of data to build their models. In \cite{snell2017prototypical}, a non-linear embedding is learned in an end-to-end manner. By minimizing the distance between different feature codes and their label-dependent central points, a non-linear classifier is constructed. 

Siamese network \cite{koch2015siamese} uses a shared feature extractor to generate feature embeddings for both the support and query images. Based on this, relation networks \cite{sung2018learning} parameterizes and learns the classification metric using a Multi-Layer Perceptron (MLP). In \cite{yosinski2014transferable}, task-specific support images are used to fine-tune the feature extractor network. In \cite{bateni2020improved}, the authors demonstrate that a simple class-covariance-based distance metric, namely the Mahalanobis distance, leads to a significant performance improvement.

In our work, though AirFi can give decent performances without any CSI data from the new environment settings, we found that by adding on the few-shot learning techniques the performance can be further improved with a few CSI data from the deployed environment setting.

\begin{figure*}[h]
	\centering
	\includegraphics[width=16cm,height=9cm]{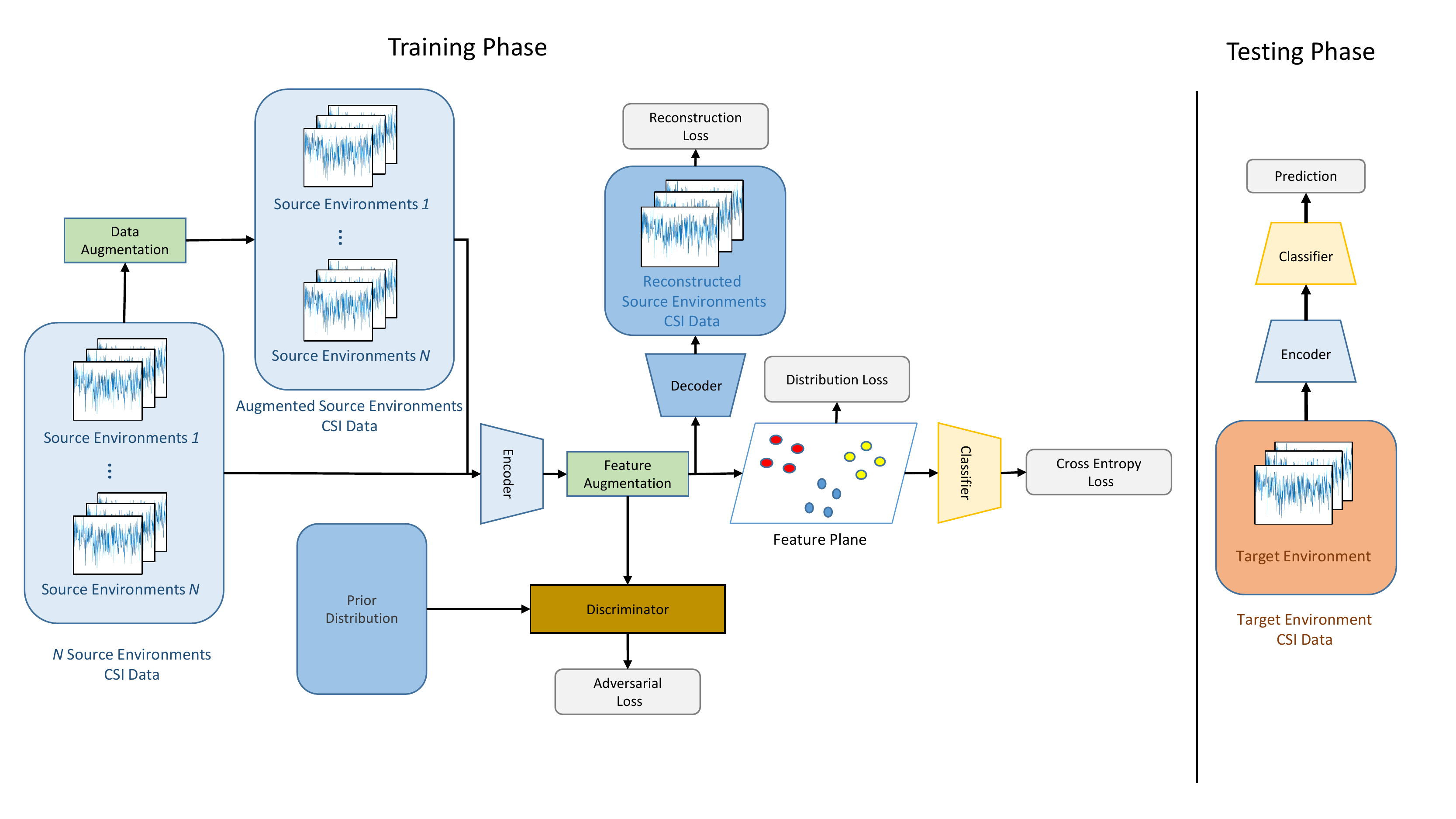}
	\caption{The structure of the proposed AirFi system. The AirFi learns robust CSI features from multiple environments by adversarial learning, feature augmentation, data augmentation and MMD alignment, and then enables the recognition model to generalize well on unseen environments.}
	\label{fig:systemchart}
\end{figure*}

\section{System Overview}

We illustrate our system AirFi in Fig \ref{fig:systemchart}. AirFi is composed of four stages: data augmentation, feature extraction \& augmentation, domain generalization and classifier training. In order to generalize the trained model, a basic assumption is that there exists a feature space underlying different domains. In \cite{wang2021multimodal}, it is proved that there is a common feature space between CSI samples of different human behavior from different environment settings. As shown in Fig \ref{fig:systemchart}, we collect data from $N$ different environments which are referred to as source environments. Then we augment our collected CSI samples by adding an arbitrary Gaussian noise, which helps to generate more simulated CSI samples. AirFi uses an encoder to encode the collected data and simulated data and extract the down-sampled features. The extracted feature codes are also augmented to improve their diversity and projected onto the hidden codes space for further training. To avoid the issue of overfitting, we take the advantage of Adversarial Autoencoders \cite{makhzani2015adversarial}. We introduce a prior distribution to regularize the distribution of the feature codes using an adversarial training approach. The decoder is used to decode the feature codes back to source environment CSI data, which helps unify the extracted feature codes. The feature codes are mapped onto the feature space underlying all source environments. AirFi minimizes the distribution variance among different training environments based on the MMD \cite{li2018domain}. Finally, a classifier is trained to recognize different human gestures using the feature codes and their corresponding label information. We will present each part of AirFi in detail in the following sections.

\subsection{Data Augmentation}

CSI samples of human gestures are collected from $N$ source environments where CSI data are relatively easier to acquire. These environments are referred to as training environments. To have more representative CSI data collections for a better generalization result, data from more domains should be collected if possible. However, due to limited time and human resources, it is very difficult and expensive to collect CSI data from all different environment settings. Furthermore, each environment is generally dynamic as well. In some previous works \cite{wang2021multimodal,xiao2019csigan,zheng2016improving}, Gaussian noise is widely used for data augmentation purpose. In \cite{wang2021multimodal}, they build a multimodal CSI model for simulated CSI data generation. They found that by adding an arbitrary Gaussian noise to the collected CSI data, they are able to generate fake CSI data. The introduced Gaussian noise will not change the label of the CSI data, moreover, these fake CSI data can be used to approximate the distribution of the related CSI data in other environment settings. 

We denote the collected CSI data pairs from $N$ different environments as $(X^N,Y)$ where $X$ is the collection of CSI samples and $Y$ is the collection of gesture labels. To augment the datasets and improve its diversity, an arbitrary Gaussian noise is added to each CSI data sequence. By combining the original collected CSI data sequence and the Gaussian noise, a new simulated CSI data sequence is generated. As explained above, the introduction of Gaussian noise does not change the label of the CSI data. As in wireless signal transmission perspective, the received signal can be modeled as a combination of the transmitted signal multiplying the transmission channel matrix and the Gaussian noise. By combining Gaussian noise with the CSI data, it will not change the class label of the data in terms of gesture recognition purposes \cite{wang2021multimodal,xiao2019csigan,huang2018multimodal}. The new generated CSI datasets can help us to approximate the CSI data distribution in other environment settings, which improves the diversity of our CSI datasets and benefits the system training.

\subsection{Feature Extraction via Adversarial Learning}

Given the augmented CSI data, an encoder is used for feature extraction. For CSI data $(X^N,Y)$ from $N$ source environments, AirFi uses the same encoder $Q$ to extract feature codes $Z$ from them. The encoder is an adversarial autoencoder. When the input CSI data pass through each convolutional layer in the encoder, they are downsampled and feature codes are extracted. 

As AirFi utilizes CSI data from all source environments to train its model, it may cause the overfitting issue during the training phase. The model trained may follow too closely to the given training data and has a strong dependency on the source environment CSI datasets. This will harm model generalization. Unless this issue is addressed, the training process of AirFi would be like a supervised learning with all source domains. It is important that the model is trained using all source domain data, meanwhile it does not have a strong dependency on the training data. Only in this way can the model learn the common feature space of CSI data from different environments and be generalized to other unseen environments. To address this issue, a prior distribution $H$ is imposed as an additional domain besides the data source environment domains. Whenever the feature codes are extracted, a regularization code is also generated from the prior distribution. Both of them are sent to the discriminator $D$ that is used to distinguish between the feature code and regularization code. This process is similar to the generative adversarial network. The adversarial loss $L_{ad}$ is given by
\begin{equation}
	L_{ad}=E_{h\sim p(h)}[\log D(h)]+E_{x\sim p(x)}[\log(1-D(Q(x)))].
\end{equation}

By minimizing the adversarial loss, we impose the dependency of the feature codes extraction on the prior distribution. This can reduce the dependency of the system model on the training CSI data. With the introduction of the prior distribution,  we expect that the issue of overfitting to the source domains data can be addressed. In theory, the prior distribution can be any arbitrary distribution \cite{li2018domain}. It is introduced to enable the adversarial encoder to extract feature code with less dependency on original CSI data distribution. Therefore the trained model can generalize better to the testing environment. We use the Laplace distribution as the prior distritbution in AirFi. We compare between several popular distribution used in related works \cite{li2018domain} and \cite{goodfellow2014generative}, such as Laplace distribution, Gaussian distribution and Uniform distribution, and the Laplace distribution performs the best. Besides, a decoder $P$ is also applied to decode the feature code back to the source domain CSI data form. The reconstruction loss $L_{re}$ is given by

\begin{equation}
	L_{re}=\sum_{n=1}^{N}\Vert P(z)_n-X_n\Vert_2^2
\end{equation}

The reconstruction process can unify the content of feature codes encoded from different training environments and improve the level of generalization to other unseen environments.

\subsection{Label Dependent Feature Augmentation}

To further enhance the generalization ability of AirFi, we augment the feature codes. As shown in \cite{li2021simple}, besides data augmentation, feature augmentation is also able to improve the model generalization by improving the diversity of feature codes. Given the collection of feature code  $Z$, which is extracted from the input CSI data $X$ using the encoder $Q$. We have

\begin{equation}
 z=Q(x).
\end{equation}

Then we input the feature codes into the augmentation layers $A$. In \cite{li2021simple}, the feature codes are augmented by scaling and adding bias in their networks. The scaling changes the absolute difference between elements in the feature codes, while the bias changes the absolute mean value of the feature codes. In AirFi, after the feature extractor, the sampled CSI feature codes $z$ are multiplied and added with random variables sampled from normal distributions. The collection of augmented feature codes is denoted as $Z'$ given by

\begin{equation}
	z'=A(z)=\alpha*z+\beta,
\end{equation}
where $\alpha$ and $\beta$ are the scale and bias hyperparameters, and sampled from two Gaussian distributions,
\begin{equation}
	\alpha\sim N(1,\sigma_1),	
\end{equation} 

\begin{equation}
	\beta\sim N(0,\sigma_2),
\end{equation} 
where $\sigma_1$ and $\sigma_2$ are two scalar hyperparameters. We set $\sigma_1=\sigma_2$ to reduce the number of hyperparameters. The perturbation introduced improves the diversity of feature codes and benefits the model generalization. However, one of the limitations brought by the feature augmentation is that the perturbation caused by the random noise may not follow the class-preserving direction \cite{li2021simple,li2021metasaug,li2021transferable,wang2022generalizing}. The augmented feature codes may lose some properties of their own behavior classes and are embedded with some new properties of other behavior classes. This will affect the model training and performances. In order to augment the feature codes to improve their diversity meanwhile preserving those feature properties of their own behavior classes, we add a label dependent regularizer $\epsilon$ to the augmentation layer in AirFi. The augmentation process becomes

\begin{equation}
	z'=A(z)=\alpha*z+\beta+\epsilon.
\end{equation}

The regularizer $\epsilon$ is sampled from a class-wise normal distribution $N(0,\Sigma_c)$, $c\in[1,C]$, where $c$ is the gesture class index, $C$ is the total number of gesture classes and $\Sigma_c$ is the class-wise covariance. $\Sigma_c$ is estimated and updated from every mini-batch of training data in a moving average manner

\begin{equation}
	\Sigma_c=\lambda*\Sigma_c+(1-\lambda)*Cov(z'|y=c),
\end{equation}
where $\lambda$ is the discount factor. The corresponding $\Sigma_c$ is only updated when the label $y$ of CSI data belongs to its own class $c$. During the training phase, AirFi calculates the class-wise covariances of data from each class and update the $\Sigma_c$ of each class. Then the elements of $\epsilon$ are sampled as,

\begin{equation}
	\epsilon_c\sim N (0,\Sigma_c). 
\end{equation}

With the additional regularizer, the feature codes are augmented more aggressively along the cross domain direction instead of the class-wise direction. Though the feature code is augmented with random variables, it is added with the covariance $\epsilon_c$ of its own gesture class $c$ to preserve key properties of that particular class. As a result the augmented feature codes are similar to those original feature codes of their own classes and have some perturbation introduced by the random variables $\alpha$ and $\beta$. This improves the diversity of feature codes from different gesture classes and leads to better performances of AirFi. We perform an ablation study to test it in our experiments.

\subsection{Domain Generalization}

The feature codes are mapped onto the feature space for domain generalization. Denote the augmented feature codes $z'$ from $n_{th}$ source environment as $Z_n$ with the distribution $\mathcal{P}_n$ of CSI data. To perform the mapping, a mean map operation $\mu(\cdot)$ is required to map the feature codes to a reproducing kernel Hilbert space \cite{smola2007hilbert}, which is given as

\begin{equation}
	\mu_{P}=E_{z\sim \mathcal{P}}[k(z')],
\end{equation}
where $k$ is the kernel function. For AirFi, we use the Radial Basis Function (RBF) kernel, which is a well-known and commonly used characteristic kernel \cite{li2018domain}. 

To achieve domain generalization of the system model, the mapped feature codes from different domains are supposed to be clustered together on the feature space. AirFi fulfills this purpose by minimizing the MMD between different distributions. The MMD between feature codes from two source environments can be measured by

\begin{equation}
MMD(Z_i,Z_j)=\Vert \mu_{\mathcal{P}_i}-\mu_{\mathcal{P}_j}\Vert.
\end{equation}

By extending it from two source environments to multiple environments, the distribution variances between different feature domains are calculated as as

\begin{equation}
	\begin{aligned}
		\frac{1}{N}\sum_{i=1}^{N}\Vert \mu_{\mathcal{P}_i}-\mu_{\mathcal{P}}\Vert &= \frac{1}{N}\sum_{i=1}^{N}\Vert \mu_{\mathcal{P}_i}-\frac{1}{N}\sum_{j=1}^{N}\mu_{\mathcal{P}_j}\Vert\\
		&=\frac{1}{N}\sum_{i=1}^{N}\Vert\sum_{j=1}^{N}\frac{1}{N}(\mu_{\mathcal{P}_i}-\mu_{\mathcal{P}_j})\\
		&\leqslant\frac{1}{N^2}\sum_{1\leqslant i,j \leqslant N}MMD(Z_i, Z_j),
	\end{aligned}
\end{equation}
where $\mu_{\mathcal{P}}$ is the mean distribution for all training environments. The indices of any two source environments are denoted by $i$ and $j$. As shown in the equation, the distribution variances are upper bounded. Therefore we use the distribution regularization loss $L_{MMD}$

\begin{equation}
L_{MMD}(Z_1,...,Z_n)=\frac{1}{N^2}\sum_{1\leqslant i,j \leqslant N}MMD(Z_i, Z_j).
\end{equation}

By minimizing $L_{MMD}$, the distribution variances between each source domain are also reduced. As a result, the model trained can be generalized between different domains.

\subsection{Classifier Optimization}

With the features identified, a classifier is added at the end of AirFi. The classifier consists of three fully connected layers $F$. With the generalized feature codes on the feature space and corresponding gesture labels, AirFi trains its classifier in a supervised learning manner. AirFi uses the cross-entropy loss to measure classification errors $L_{ce}$:

\begin{equation}
		L_{ce}= -E_{z,y}\log[p(y|z)].
\end{equation}

\subsection{Few-Shot Learning}

In our study, we find that though AirFi can achieve a decent performance without any training CSI samples from the target environments, a few CSI samples from the target environment can indeed help to further improve its performances. It is possible that the feature codes encoded from testing environment CSI data may not be mapped closely to the distribution of source environment feature codes. To address this issue, a few labeled CSI samples from the testing environment can be very helpful. After AirFi is trained using the source domain data from the training environment, we input the testing environment CSI data and minimize the distribution difference between the source environment and testing environment CSI data with the new distribution regularizer loss $L'_{MMD}$. 

\begin{equation}
	L'_{MMD}=MMD(Z_{training},Z_{testing})
\end{equation}

We retrain the system with the source environment, target environment distribution loss and classification cross-entropy loss $L'_{ce}$ which also includes the testing environment labeled data $X_t$.

\begin{equation}
	L'_{ce}= -E_{x_t,y_t}\log[p(y_t|x_t)].
\end{equation}

Then the total few shot learning loss $L_f$ is given by

\begin{equation}
	L_f=L'_{MMD} + L'_{ce}
\end{equation}

 After the few-shot learning is added, the trained model has a better generalization ability on the target environment. We show the improvement in Section IV.

\begin{figure*}[htb]
	\centering
	\subfigure[Lab layout]{
		\begin{minipage}[t]{0.5\linewidth}
			\centering
			\includegraphics[width=7.5cm,height=5cm]{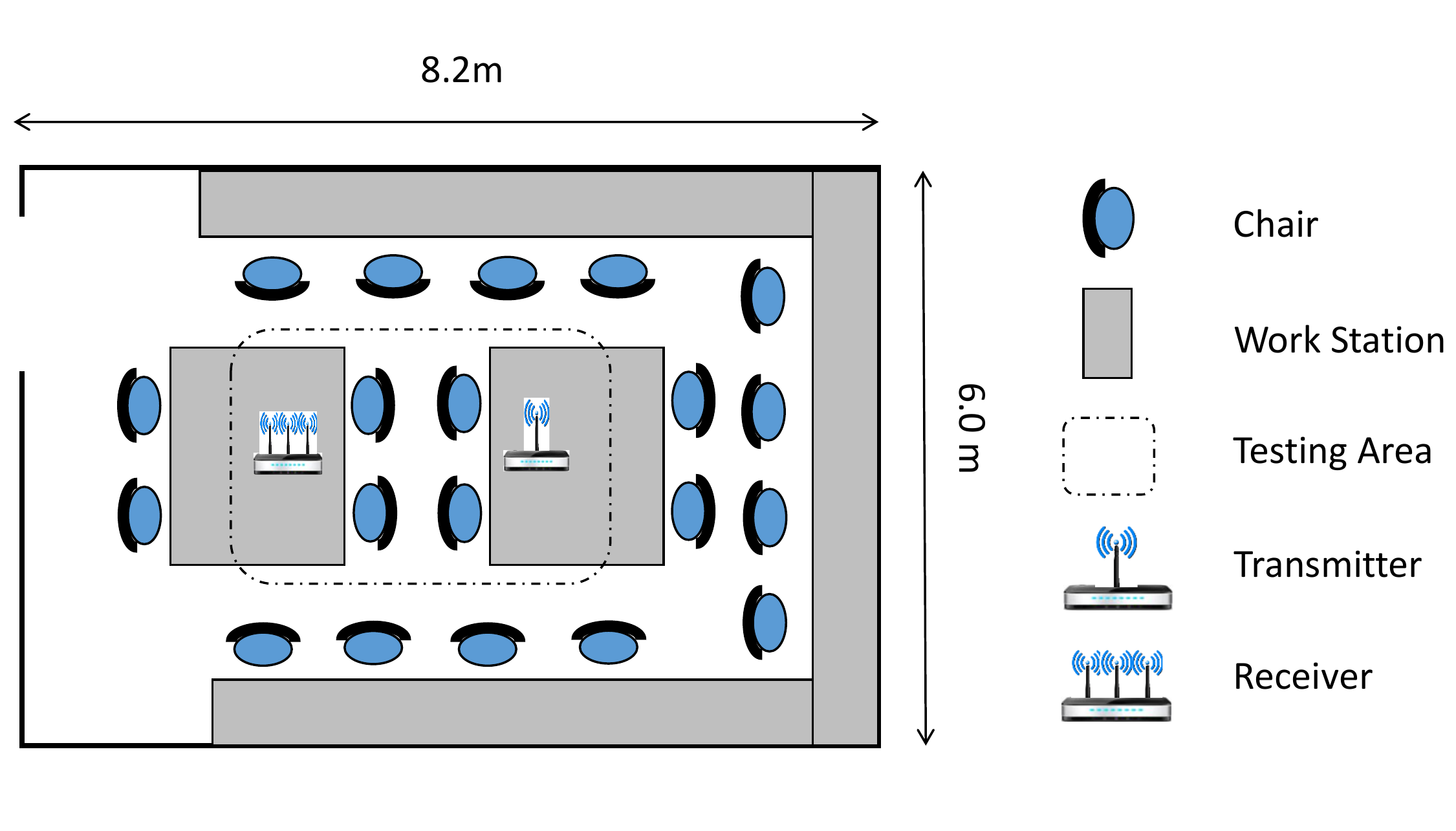}
		\end{minipage}%
	}%
	\subfigure[Cubic office layout]{
		\begin{minipage}[t]{0.4\linewidth}
			\centering
			\includegraphics[width=7.5cm,height=5cm]{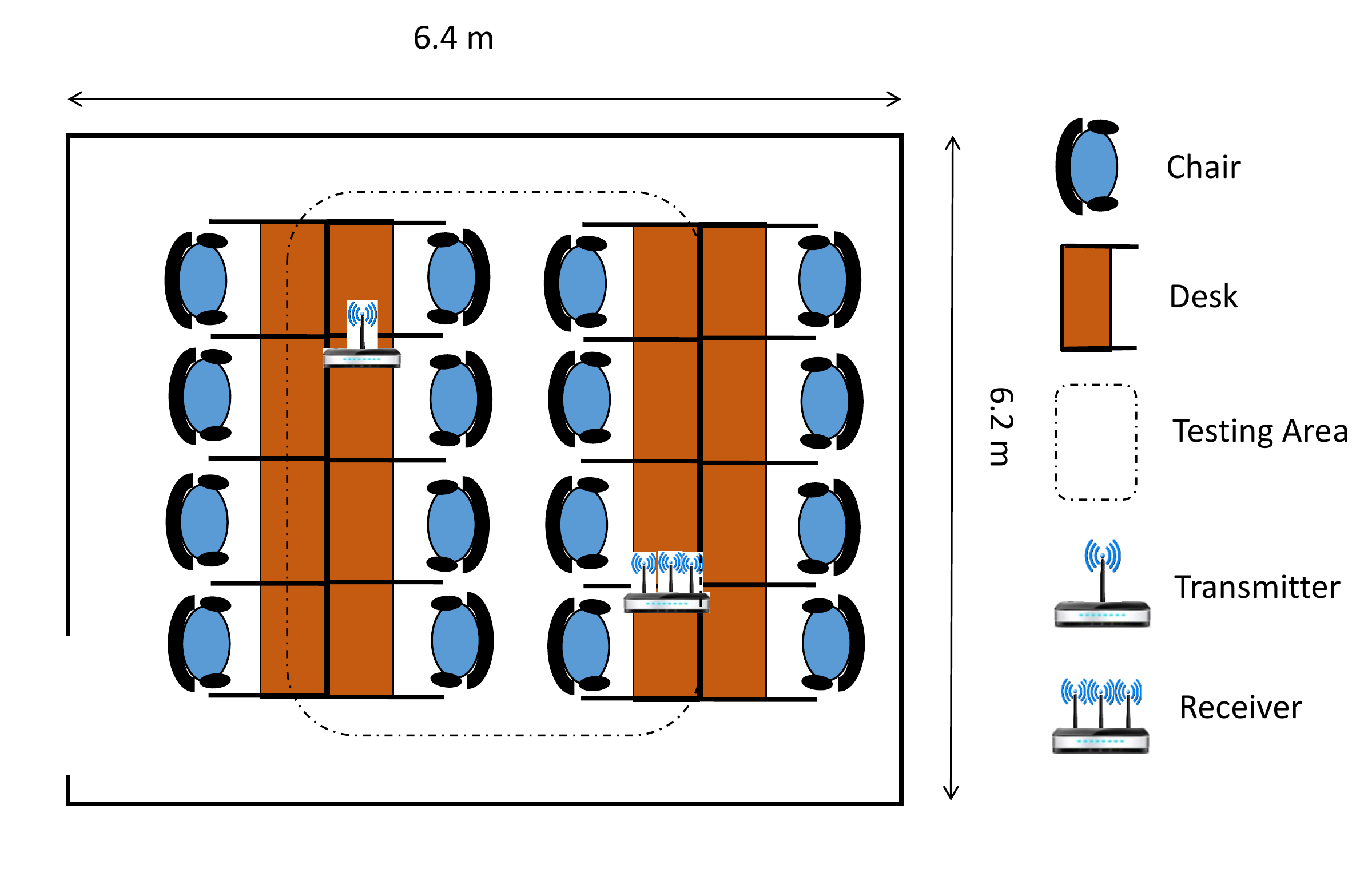}
		\end{minipage}%
	}%
	
	\subfigure[Tutorial room layout]{
		\begin{minipage}[t]{0.5\linewidth}
			\centering
			\includegraphics[width=7.5cm,height=5cm]{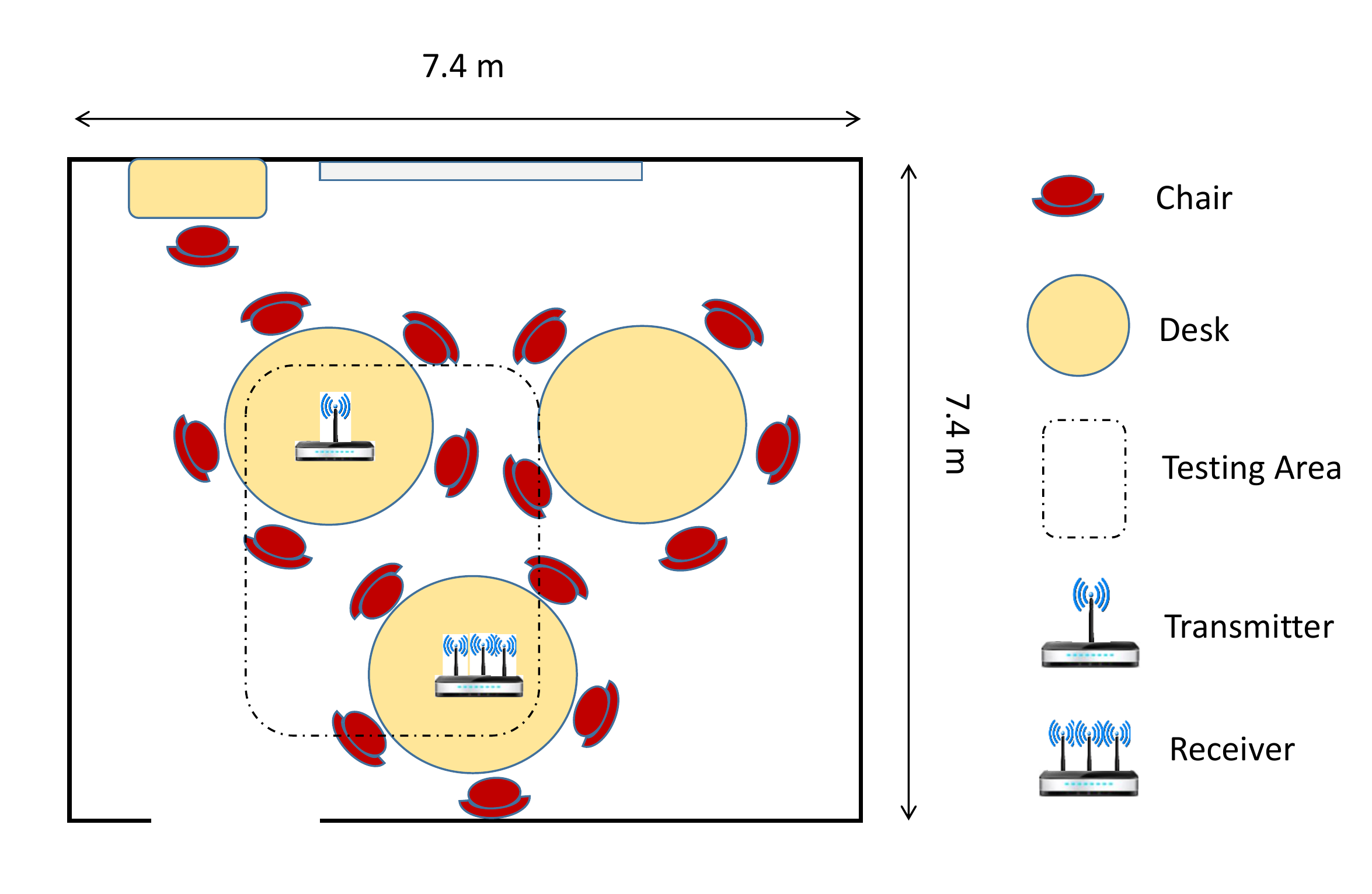}
		\end{minipage}%
	}%
	\subfigure[Meeting room layout]{
		\begin{minipage}[t]{0.4\linewidth}
			\centering
			\includegraphics[width=7.5cm,height=5cm]{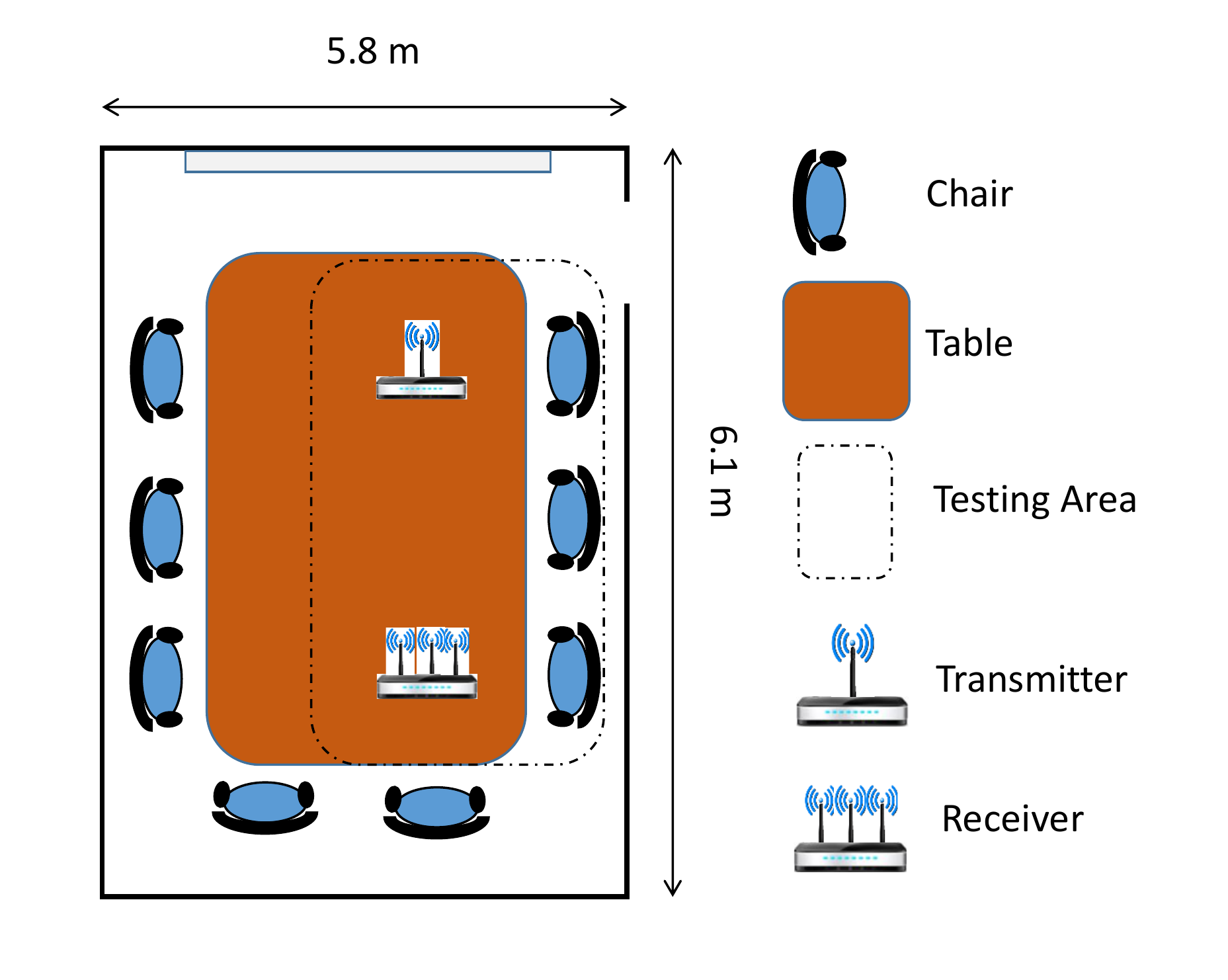}
		\end{minipage}%
	}%
	\centering
	\caption{Layouts of experimental environment}
	\label{fig:layout}
\end{figure*}

\begin{figure*}[htb]
	\centering
	\subfigure[Circle Arm]{
		\begin{minipage}[t]{0.25\linewidth}
			\centering
			\includegraphics[width=4.6cm,height=3cm]{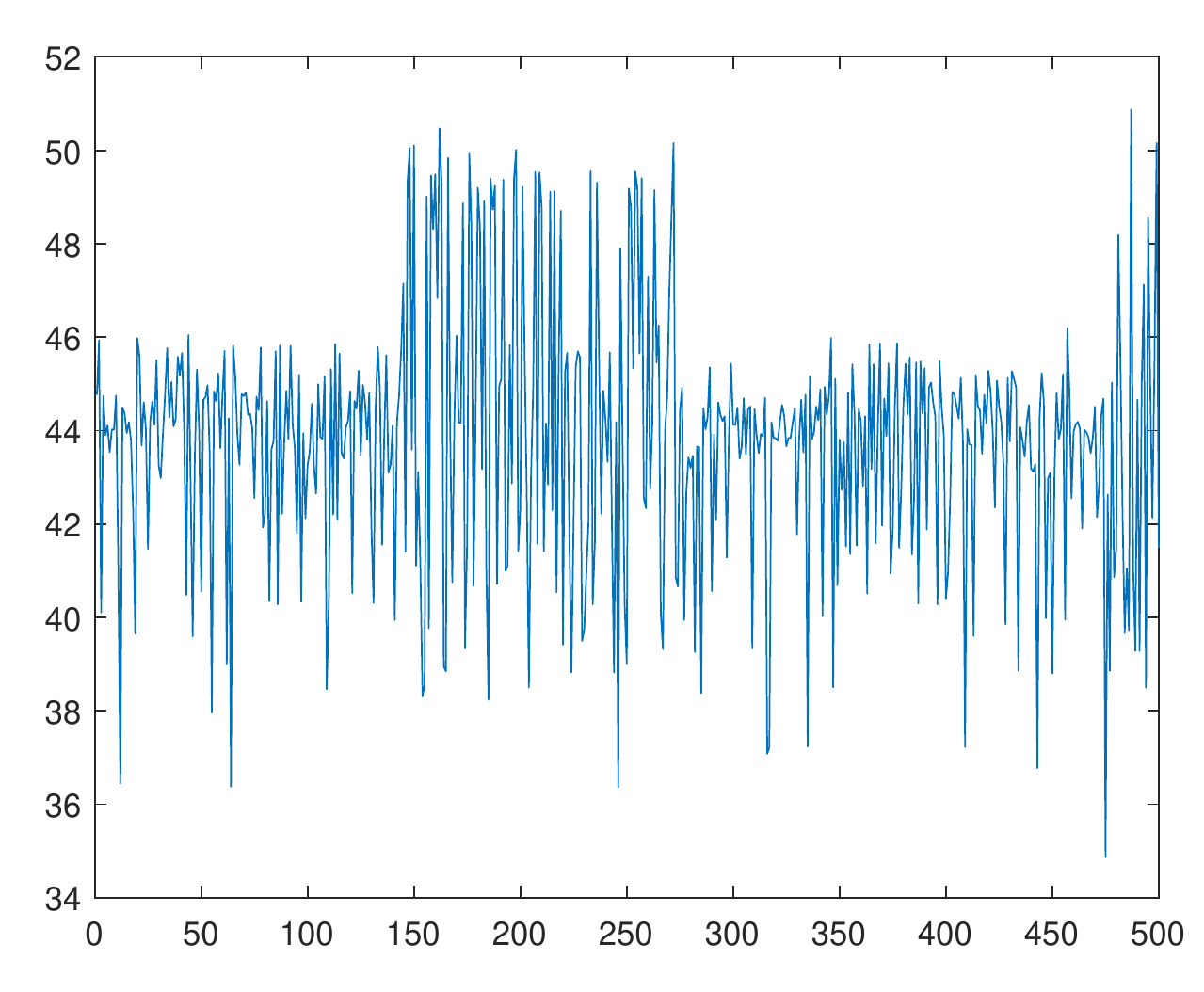}
		\end{minipage}%
	}%
	\subfigure[Clap]{
		\begin{minipage}[t]{0.25\linewidth}
			\centering
			\includegraphics[width=4.6cm,height=3cm]{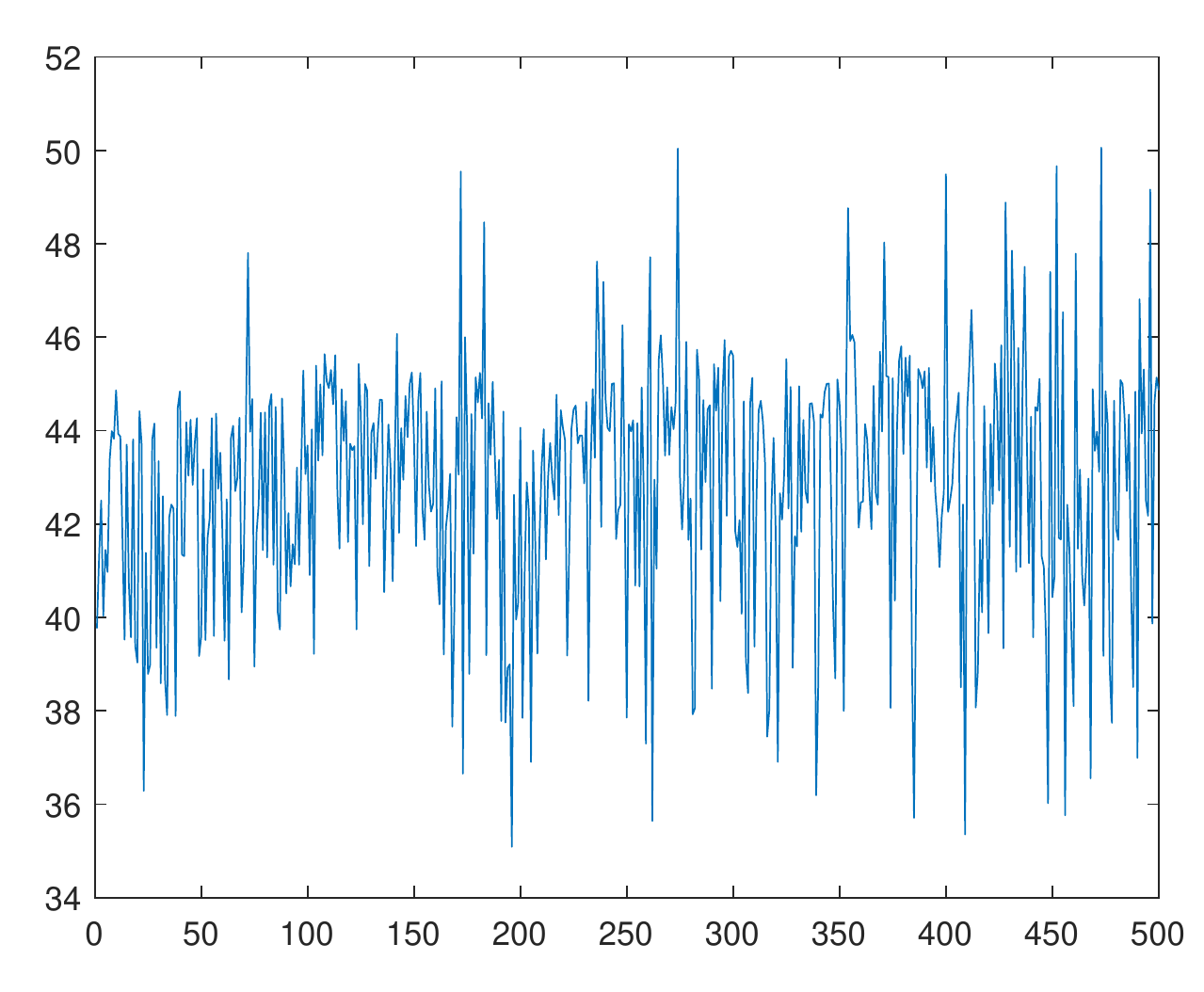}
		\end{minipage}%
	}%
	\subfigure[Front and Back]{
		\begin{minipage}[t]{0.25\linewidth}
			\centering
			\includegraphics[width=4.6cm,height=3cm]{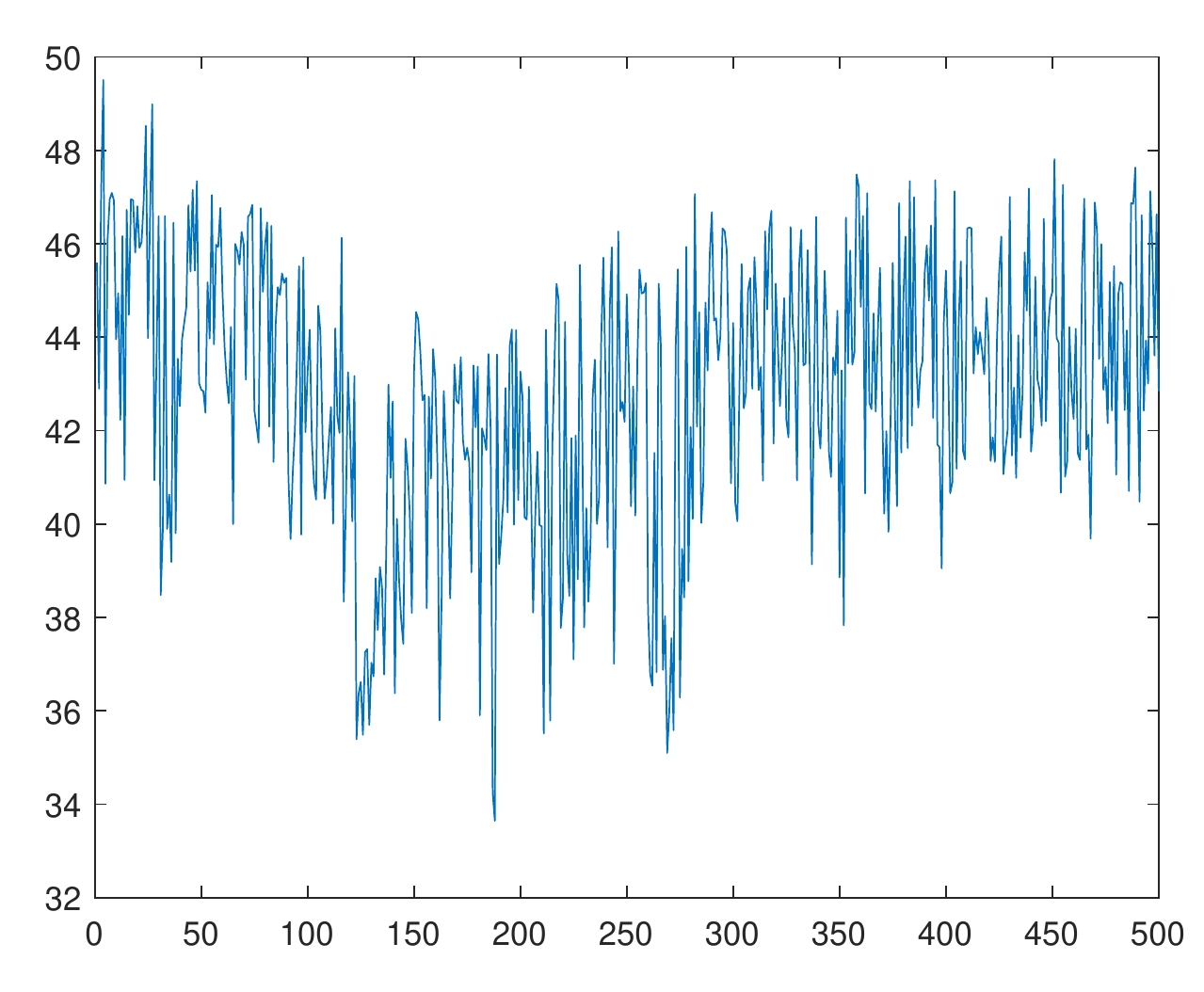}
		\end{minipage}
	}%
	\subfigure[Zoom]{
		\begin{minipage}[t]{0.25\linewidth}
			\centering
			\includegraphics[width=4.6cm,height=3cm]{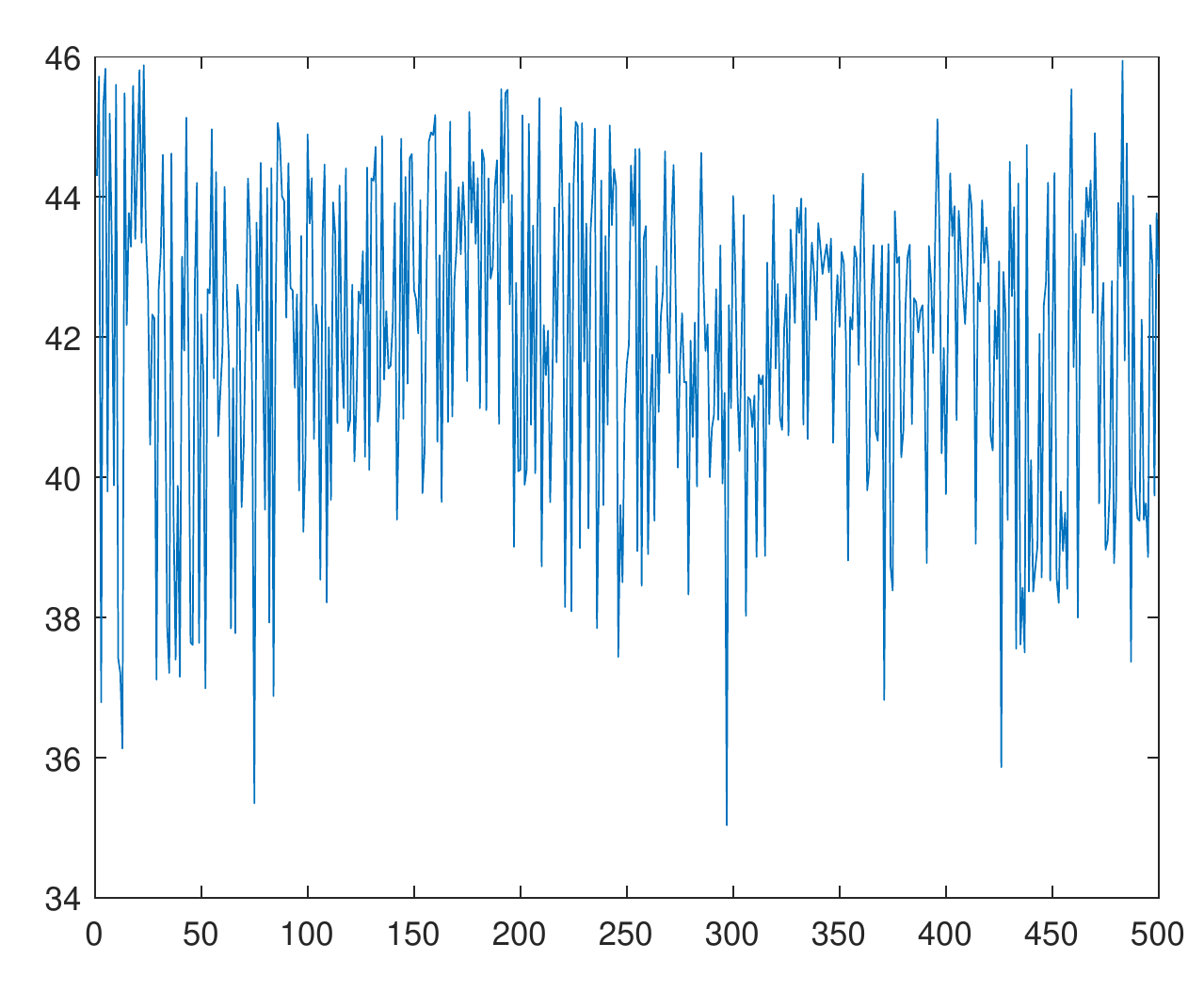}
		\end{minipage}
	}%
	
	\subfigure[Fist]{
		\begin{minipage}[t]{0.25\linewidth}
			\centering
			\includegraphics[width=4.6cm,height=3cm]{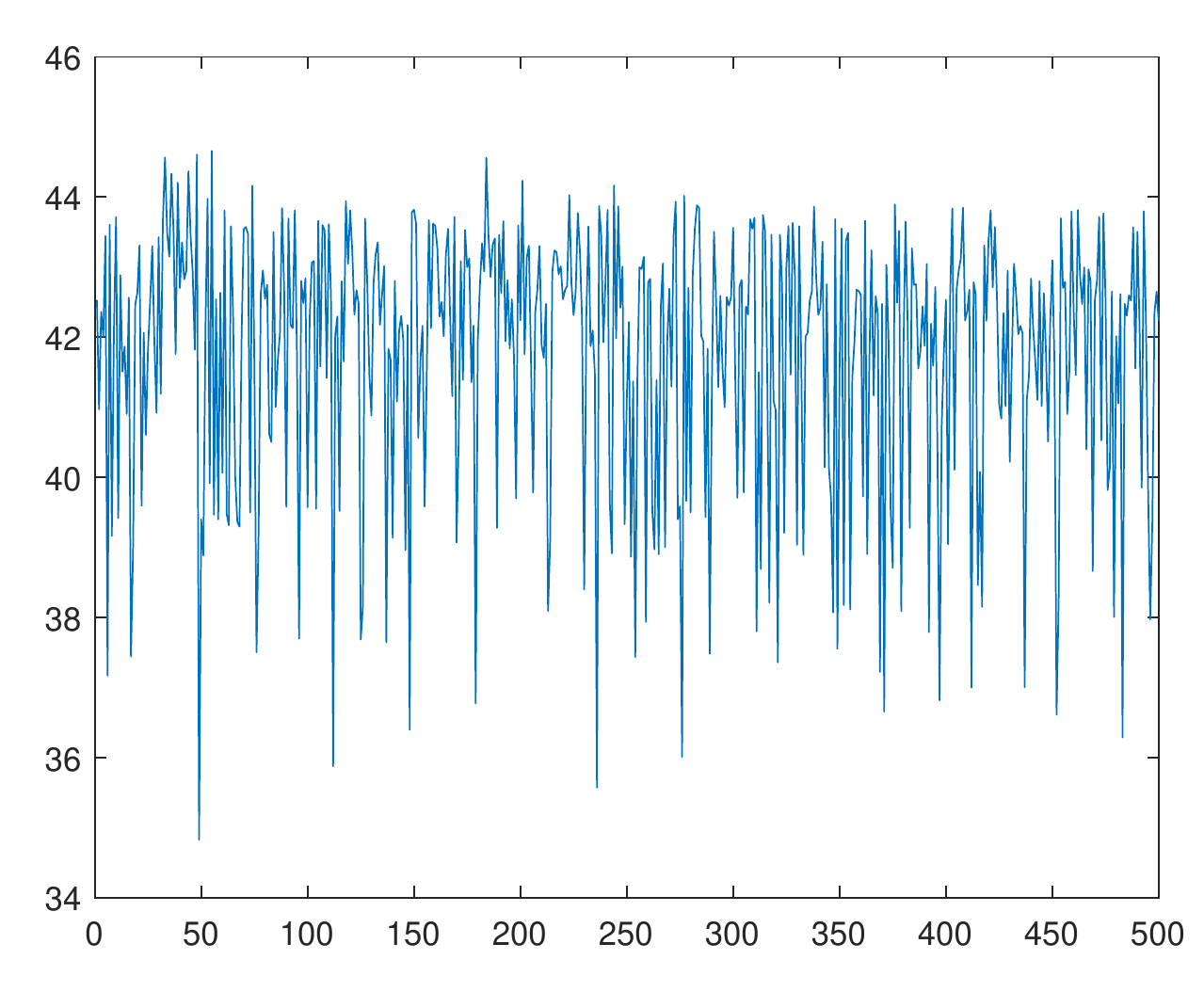}
		\end{minipage}%
	}%
	\subfigure[Left and Right]{
		\begin{minipage}[t]{0.25\linewidth}
			\centering
			\includegraphics[width=4.6cm,height=3cm]{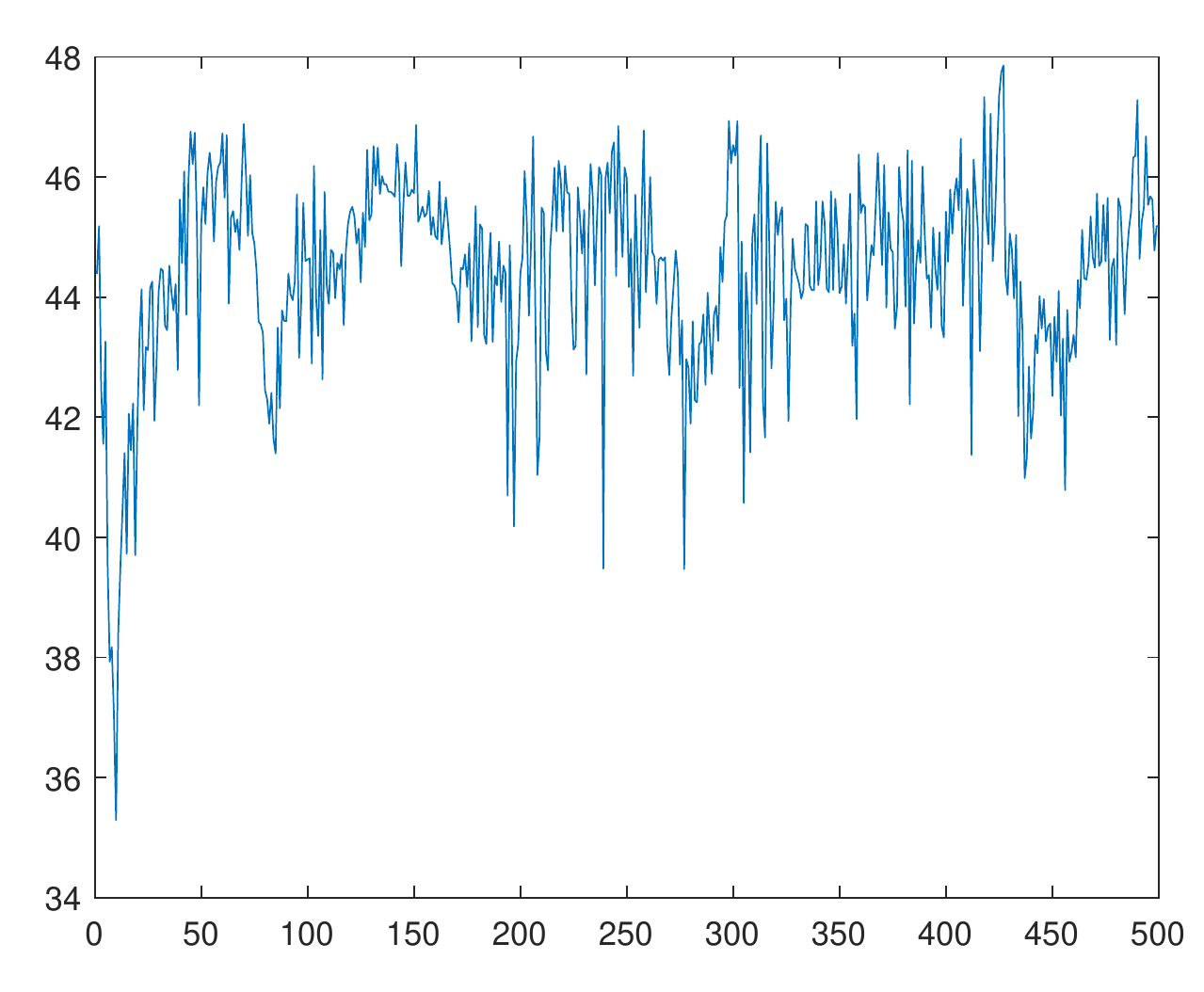}
		\end{minipage}%
	}%
	\subfigure[Throw]{
		\begin{minipage}[t]{0.25\linewidth}
			\centering
			\includegraphics[width=4.6cm,height=3cm]{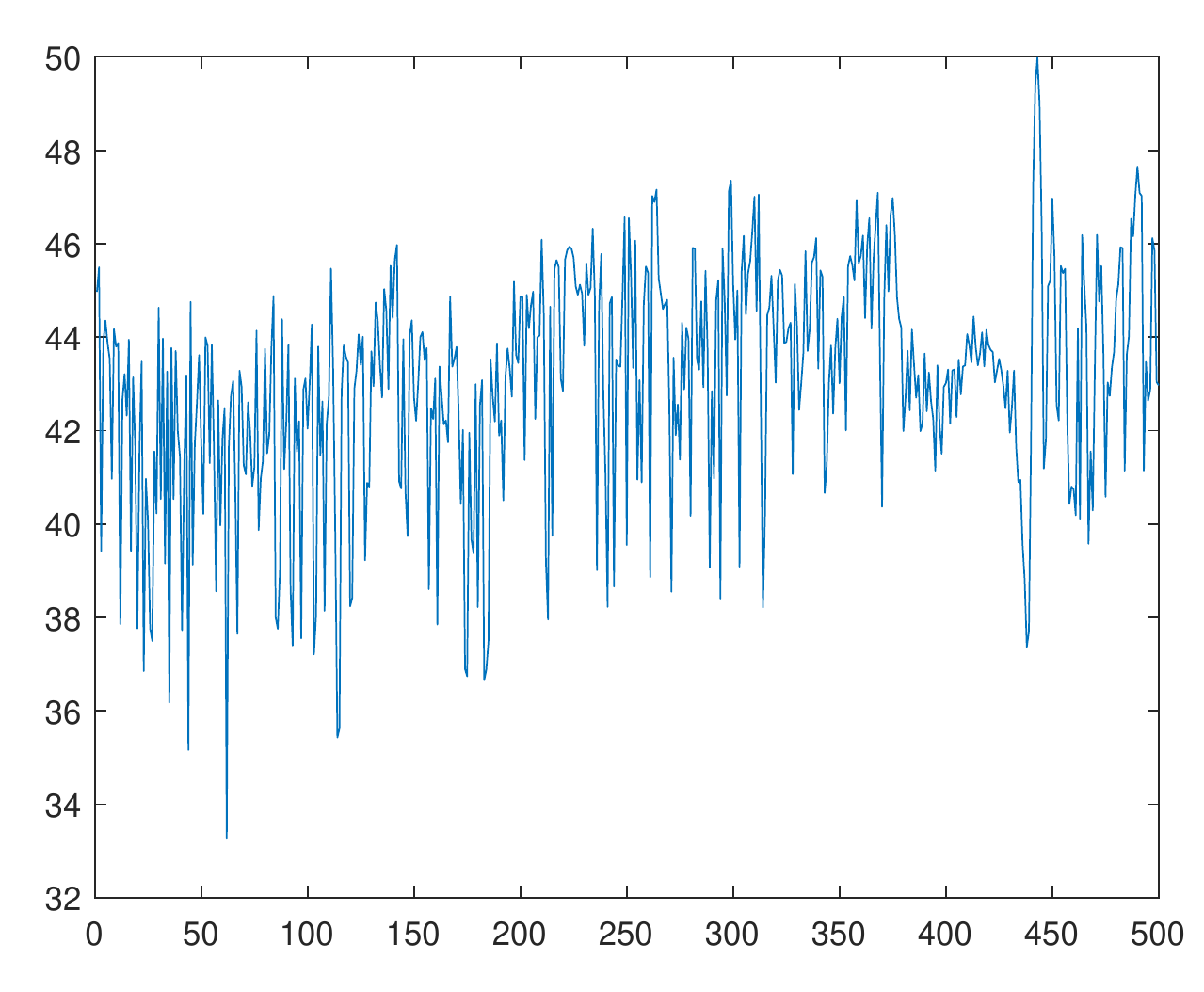}
		\end{minipage}
	}%
	\subfigure[Up and Down]{
		\begin{minipage}[t]{0.25\linewidth}
			\centering
			\includegraphics[width=4.6cm,height=3cm]{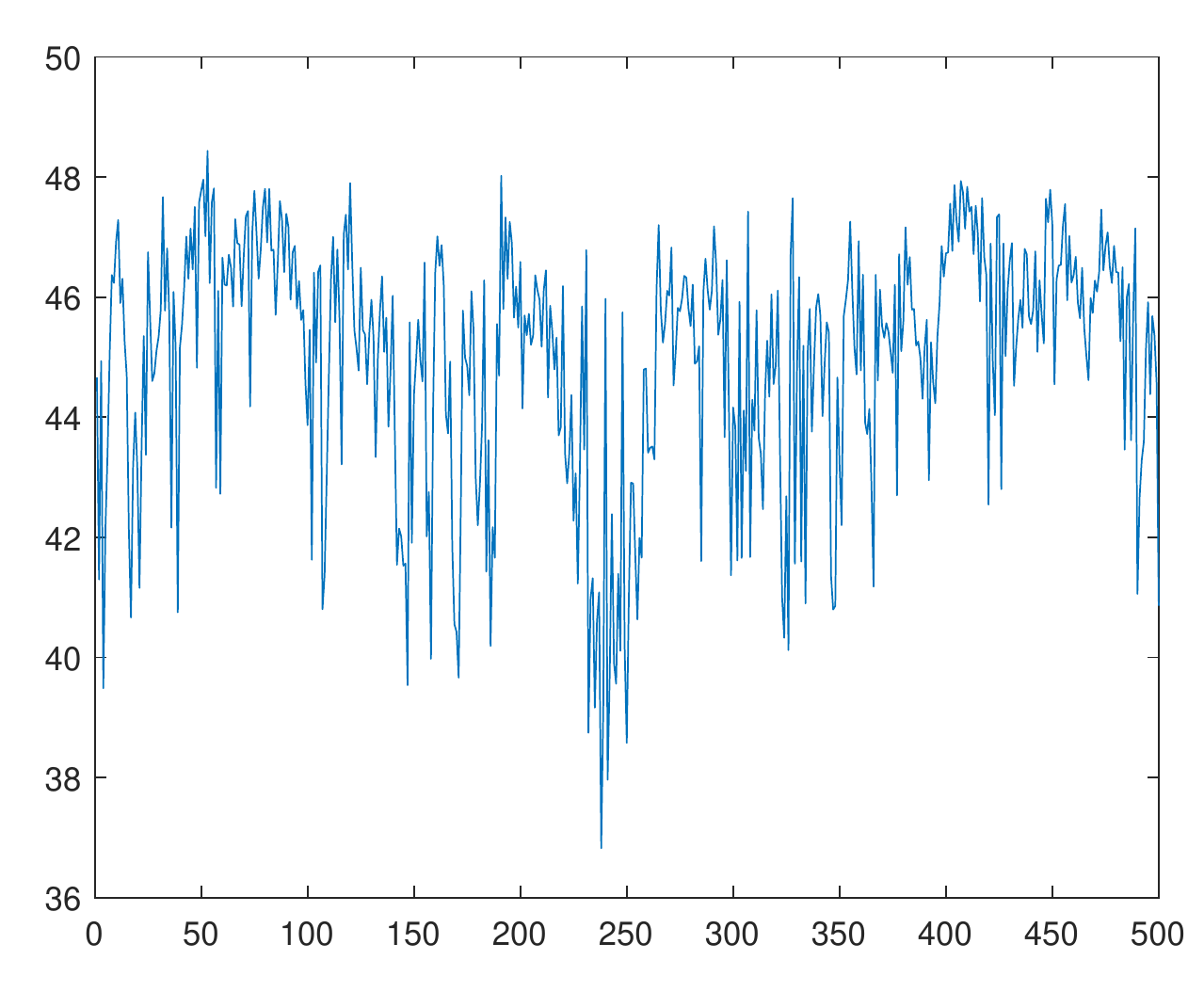}
		\end{minipage}
	}%
	
	\centering
	\caption{CSI plots of different human gestures}
	\label{fig:csiplot}
\end{figure*}

\begin{table*}[htb]
	\begin{center}
		\caption{Overall Performances Evaluation}
		\begin{tabular}{c|c|c|c|c|c|c|c|c|c}
			\toprule
			\textbf{Sytems}  &\textbf{Up\&Down}  & \textbf{Left\&Right} &  \textbf{Front\&Back} & \textbf{Clap} & \textbf{Fist}& \textbf{Waving}& \textbf{Throw}& \textbf{Zoom}& \textbf{Environment Index } \\
			\midrule
			{\textbf{CNN Only \cite{wang2021multimodal}}}  &31.23\% & 29.57\%  & 36.55\%  & 32.48\%  &29.63\%  &34.11\%  &30.86\%  &28.99\%  &\\
			{\textbf{WiGr \cite{zhang2021wifi}}} &80.67\% & 81.44\% & 79.38\%   &  82.04\%  &80.78\%  &78.63\%  &79.81\%  &80.52\%  &  \\
			{\textbf{WGRDTL \cite{bu2018wi}}}  &70.85\% &73.93\%  & 73.47\%  & 74.01\% &72.14\%  &70.65\%  &71.46\%  &69.19\%  & ABC-D\\
			{\textbf{Wi-Multi \cite{feng2019wi}}}& 72.15\%  &71.89\%  & 69.94\%  & 71.28\%  &70.58\%  &72.23\%  &70.18\%  &70.63\%  &  \\
			{\textbf{AirFi}}  &\textbf{89.32\%} & \textbf{91.02\%}  & \textbf{89.47\%}  & \textbf{91.66\%}  & \textbf{90.85\%}  &\textbf{88.79\%} & \textbf{92.37\%}  & \textbf{89.61\%}  &\\
			\midrule
			{\textbf{CNN Only \cite{wang2021multimodal}}}  & 35.95\% & 39.11\%  & 37.49\%  & 46.24\%  & 40.18\%  & 41.79\%  &39.85\%  &40.43\%  &\\
			{\textbf{WiGr \cite{zhang2021wifi}}} &76.39\% & 79.20\% & 80.71\%   & 78.58\%  & 81.84\%  & 79.98\%  & 79.34\%  &  80.86\%&  \\
			{\textbf{WGRDTL \cite{bu2018wi}}}  &73.97\% &73.55\%  & 70.11\%  & 73.16\%  & 71.74\%  & 70.06\%  &70.21\%  &72.63\%  & ABD-C \\
			{\textbf{Wi-Multi \cite{feng2019wi}}}& 74.14\%  & 71.08\%  & 70.83\%  & 70.92\%  & 69.71\%  & 70.57\%  &69.24\%  &73.26\%  &\\
			{\textbf{AirFi}}  &\textbf{87.48\%} &\textbf{91.25\%}  & \textbf{91.28\%}  & \textbf{90.75\%}  &\textbf{92.02\%}  &\textbf{90.79\%}  & \textbf{89.31\%}  & \textbf{90.63\%}  &\\
			\midrule
			{\textbf{CNN Only \cite{wang2021multimodal}}}  & 31.01\% & 40.42\%  & 34.15\%  & 32.68\%  & 36.76\%  & 37.91\%  & 35.31\%  &37.82\%  &\\
			{\textbf{WiGr \cite{zhang2021wifi}}} & 79.38\% & 82.57\% & 83.61\%   & 79.99\%  & 82.52\%  & 81.64\%  & 77.24\%  & 80.36\%  &\\
			{\textbf{WGRDTL \cite{bu2018wi}}}  &71.13\% & 74.10\%  & 76.97\%  & 74.45\%  & 75.23\%  &72.28\%  &70.34\%  &70.46\%  & ACD-B  \\
			{\textbf{Wi-Multi \cite{feng2019wi}}}& 73.69\%  & 71.25\%  & 70.98\%  & 72.89\%  & 74.35\%  & 72.47\%  & 74.84\%  &71.06\%  & \\
			{\textbf{AirFi}}  &\textbf{88.16\%} & \textbf{89.72\%}  & \textbf{92.78\%}  & \textbf{91.06\%}  & \textbf{90.34\%}  & \textbf{89.47\%}  & \textbf{88.32\%}  & \textbf{90.30\%}  &\\
			\midrule
			{\textbf{CNN Only \cite{wang2021multimodal}}}  &38.44\% & 37.98\%  & 32.17\%  & 30.19\%  & 31.74\%  & 32.68\%  & 32.48\%  & 36.37\%  &\\
			{\textbf{WiGr \cite{zhang2021wifi}}} &76.52\% & 79.63\% & 80.46\%   &  81.09\%  & 76.87\%  &79.68\%  &80.50\%  &80.72\%  &  \\
			{\textbf{WGRDTL \cite{bu2018wi}}}  &72.54\% & 73.71\%  & 70.38\%  & 72.02\%  & 73.44\%  & 72.18\%  & 71.51\%  & 72.43\%  & BCD-A \\
			{\textbf{Wi-Multi \cite{feng2019wi}}}& 71.84\%  & 72.93\%  & 74.42\%  & 77.59\%  & 73.28\%  & 73.80\%  & 72.17\%  & 74.67\%  &\\
			{\textbf{AirFi}}  & \textbf{91.73\%} & \textbf{90.86\%}  &\textbf{87.26\%}  & \textbf{88.67\%}  &\textbf{89.42} \%  & \textbf{89.76\%}  & \textbf{90.92}\%  & \textbf{87.58\%}  &\\
			\bottomrule
			
		\end{tabular}
		\label{tab:table1}
	\end{center}
\end{table*}

\begin{table*}[htb]
	\begin{center}
		\caption{Ablation Study}
		\begin{tabular}{c|c|c|c|c|c|c|c|c|c}
			\toprule
			\textbf{Sytems}  &\textbf{Up\&Down}  & \textbf{Left\&Right} &  \textbf{Front\&Back} & \textbf{Clap} & \textbf{Fist}& \textbf{Waving}& \textbf{Throw}& \textbf{Zoom}& \textbf{Environment Index } \\
			\midrule
			{\textbf{CNN Only}}  &31.23\% & 29.57\%  & 36.55\%  & 32.48\%  &29.63\%  &34.11\%  &30.86\%  &28.99\%  &\\
			{\textbf{AirFi w/o Data Aug}} &85.15\% & 87.06\% & 83.47\%   &  86.72\%  &85.96\%  &87.17\%  &85.63\%  &86.31\%  & ABC-D \\
			{\textbf{AirFi w/o Fea Aug}}  &84.34\% &85.65\%  & 84.98\%  & 86.10\% &83.57\%  &86.21\%  &85.67\%  &82.44\%  & \\
			{\textbf{AirFi}}  &\textbf{89.32\%} &\textbf{91.02\%}  & \textbf{89.47\%}  & \textbf{91.66\%}  &\textbf{90.85\%}  &\textbf{88.79\%}  &\textbf{92.37\%}  &\textbf{89.61\%}  &\\
			\midrule
			{\textbf{CNN Only}}  &31.23\% & 29.57\%  & 36.55\%  & 32.48\%  &29.63\%  &34.11\%  &30.86\%  &28.99\%  &\\
			{\textbf{AirFi w/o Data Aug}} &84.83\% & 86.94\% & 87.75\%   & 85.28\%  &84.67\%  &85.13\%  &83.71\%  &86.14\%  & ABD-C \\
			{\textbf{AirFi w/o Fea Aug}}  &82.68\% &85.14\%  & 84.36\%  & 86.11\% &84.09\%  &83.76\%  &82.97\%  &84.59\%  & \\
			{\textbf{AirFi}}  &\textbf{87.48\%} &\textbf{91.25\%}  & \textbf{91.28\%}  & \textbf{90.75\%}  &\textbf{92.02\%}  &\textbf{90.79\%}  & \textbf{89.31\%}  & \textbf{90.63\%}  &\\
			\midrule
			{\textbf{CNN Only}}  &31.23\% & 29.57\%  & 36.55\%  & 32.48\%  &29.63\%  &34.11\%  &30.86\%  &28.99\%  &\\
			{\textbf{AirFi w/o Data Aug}} &82.87\% & 83.94\% & 84.72\%   & 83.16\%  & 84.71\%  & 85.93\%  &83.66\%  &84.68\%  & ACD-B \\
			{\textbf{AirFi w/o Fea Aug}}  &83.91\% &82.70\%  & 83.05\%  & 84.61\% &83.14\%  &83.45\%  &83.17\%  &82.08\%  & \\
			{\textbf{AirFi}}  &\textbf{88.16\%} &\textbf{89.72\%}  & \textbf{92.78\%}  & \textbf{91.06\%}  & \textbf{90.34\%}  & \textbf{89.47\%}  & \textbf{88.32\%}  & \textbf{90.30\%}  &\\
			\midrule
			{\textbf{CNN Only}}  &31.23\% & 29.57\%  & 36.55\%  & 32.48\%  &29.63\%  &34.11\%  &30.86\%  &28.99\%  &\\
			{\textbf{AirFi w/o Data Aug}} &83.95\% & 82.27\% & 84.78\%   & 85.49\%  &82.29\%  &84.77\%  &84.52\%  &83.96\%  & BCD-A \\
			{\textbf{AirFi w/o Fea Aug}}  &83.01\% &82.94\%  & 84.95\%  & 82.39\% & 81.79\%  &83.28\%  & 83.26\%  & 84.73\%  & \\
			{\textbf{AirFi}}  & \textbf{91.73\%} & \textbf{90.86\%}  &\textbf{87.26\%}  & \textbf{88.67\%}  &\textbf{89.42\%}  & \textbf{89.76\%}  &\textbf{90.92\%}  & \textbf{87.58\%}  &\\
			\bottomrule
			
		\end{tabular}
		\label{tab:table2}
	\end{center}
\end{table*}

\begin{table*}[htb]
	\begin{center}
		\caption{Few-Shot learning test in lab}
		\begin{tabular}{c|c|c|c|c|c|c|c|c|c}
			\toprule
			\textbf{Sytems}  &\textbf{Up\&Down}  & \textbf{Left\&Right} &  \textbf{Front\&Back} & \textbf{Clap} & \textbf{Fist}& \textbf{Waving}& \textbf{Throw}& \textbf{Zoom}& \textbf{Environment Index } \\
			\midrule
			{\textbf{CNN Only \cite{wang2021multimodal}}}  &39.57\% &41.08\%  & 47.11\%  & 34.95\%  &43.89\%  &41.97\%  &43.44\%  &37.96\%  &\\
			{\textbf{WiGr \cite{zhang2021wifi}}} &86.94\% & 87.28\% & 87.43\%   &  87.62\%  & 86.15\%  &84.60\%  & 89.37\%  &90.07\%  &  \\
			{\textbf{WGRDTL \cite{bu2018wi}}}  &80.19\% &78.88\%  & 76.58\%  & 79.78\% &80.67\%  &81.85\%  &79.54\%  &78.27\%  & ABC-D \\
			{\textbf{Wi-Multi \cite{feng2019wi}}}& 78.95\%  &79.52\%  & 77.91\%  & 81.49\%  &78.18\%  &80.56\%  &80.02\%  &78.49\%  &  \\
			{\textbf{AirFi}} & \textbf{94.63\%} &\textbf{95.21\%}  & \textbf{93.55\%}  & \textbf{93.17\%}  &\textbf{91.68\%}  &\textbf{96.14\%}  &\textbf{95.24\%}  &\textbf{93.54\%} &\\
			\midrule
			{\textbf{CNN Only \cite{wang2021multimodal}}}  &45.93\% &41.08\%  & 35.12\%  & 40.58\%  &38.94\%  &44.421\%  &37.01\%  &43.18\%  &\\
			{\textbf{WiGr \cite{zhang2021wifi}}} &85.99\% & 87.19\% & 86.38\%   &  86.92\%  & 87.37\%  & 88.96\%  & 88.00\%  & 87.25\%&  \\
			{\textbf{WGRDTL \cite{bu2018wi}}}  &79.63\% &81.71\%  & 78.45\%  & 78.93\%  & 79.48\%  &80.16\%  &81.04\%  &71.67\%  & ABD-C \\
			{\textbf{Wi-Multi \cite{feng2019wi}}}& 78.59\%  &78.61\%  & 79.38\%  & 79.11\%  &  81.07\%  &77.86\%  &80.19\%  &82.93\%  &\\
			{\textbf{AirFi}}  &\textbf{93.89\%} &\textbf{93.61\%}  &\textbf{94.29\%}  &\textbf{93.28\%}  &\textbf{93.47\%}  &\textbf{93.68\%}  &\textbf{94.78\%}  &\textbf{94.13\%}  &\\
			\midrule
			{\textbf{CNN Only \cite{wang2021multimodal}}}  &48.15\% &34.18\%  & 42.69\%  & 39.48\%  &41.56\%  &42.00\%  &38.14\%  &40.82\%  &\\
			{\textbf{WiGr \cite{zhang2021wifi}}} &84.60\% & 88.47\% & 89.15\%   &  90.81\%  & 89.39\%  &90.82\%  &89.49\%  &87.30\%  & \\
			{\textbf{WGRDTL \cite{bu2018wi}}}  &81.98\% &82.19\%  & 79.58\%  & 77.14\%  &82.84\%  &80.14\%  &78.92\%  &83.86\%  &  ACD-B \\
			{\textbf{Wi-Multi \cite{feng2019wi}}}& 78.49\%  &79.18\%  & 77.39\%  & 74.81\%  & 80.18\%  &81.44\%  &78.68\%  &80.49\%  & \\
			{\textbf{AirFi}}  &\textbf{92.69\%} &\textbf{92.11\%}  &\textbf{94.27\%}  & \textbf{95.34\%}  &\textbf{93.74\%}  &\textbf{95.08\%}  &\textbf{94.18\%}  &\textbf{94.57\%}  &\\
			\midrule
			{\textbf{CNN Only \cite{wang2021multimodal}}}  &41.33\% &47.86\%  & 42.17\%  & 39.68\%  & 39.42\%  &43.71\%  &43.62\%  &44.76\%  &\\
			{\textbf{WiGr \cite{zhang2021wifi}}} &86.07\% & 86.32\% & 88.95\%   &  87.60\%  & 87.23\%  & 90.27\%  &89.08\%  & 88.81\%  &  \\
			{\textbf{WGRDTL \cite{bu2018wi}}}  &79.31\% &82.18\%  & 83.86\%  & 79.41\%  & 80.09\%  & 79.47\%  & 78.86\%  &81.10\%  & BCD-A \\
			{\textbf{Wi-Multi \cite{feng2019wi}}}& 75.89\%  & 77.21\%  & 82.12\%  & 82.63\%  & 80.76\%  & 81.97\%  & 79.55\%  & 81.83\%  &  \\
			{\textbf{AirFi}}  &\textbf{94.58\%} &\textbf{93.89\%}  &\textbf{94.79\%}  & \textbf{92.88\%}  &\textbf{93.75\%}  &\textbf{95.86\%}  &\textbf{95.33\%}  &\textbf{94.17\%}  &\\
			\bottomrule
			
		\end{tabular}
		\label{tab:table3}
	\end{center}
\end{table*}

\section{Experiment}

To be applied in a different environment, AirFi does not require any CSI data from the target environments as in the case for most existing systems. Using CSI data from several training environments, AirFi aims to build a generalized system model. In this section, we conduct multiple experiments to evaluate AirFi under different environments. Firstly, we introduce the experimental setup. Then we do an overall evaluation to compare AirFi with other CSI-based smart human sensing systems. Thirdly, we do an ablation study to investigate the impacts of different components in AirFi. Besides, we test AirFi with few-shot learning added on and observe how it improves the performances. Finally, we use the T-SNE plots to show the distribution of hidden features with different system designs.

\subsection{Environment Setup and Data collection} 

AirFi is designed to be generalizable to different environment settings. In order to test its ability, our experiments are performed in four different environment settings: lab, cubic office, meeting room and tutorial room. We select them as their layouts and furniture are very different from each other, which can be used to test the performances of compared systems in different environments. Their layouts are shown in Fig \ref{fig:layout}. In each location, two routers are used. One router is the transmitter (one antenna), and the other router is the receiver (three antennas). We have upgraded the firmware of both routers to our CSI enabled platform for data collection. The transmitter is operated in 802.11n AP mode at 5 GHz with a 40 MHz bandwidth and the receiver is connected to the transmitter in client mode. The detail of the environments is as follows.

\begin{enumerate}
	\item Environment A (lab environment). The furniture in the lab is mainly lab benches and chairs. The routers are placed on two opposite lab benches with a distance of 2.3 meters. The volunteer performs different human gestures, while sitting in the middle between the two lab benches.
	\item Environment B (cubic office environment). The furniture in the cubic office is mainly cubical desks and chairs. The two routers are placed on two different desks as shown in the layout figure with a distance of 2.3 meters. The volunteer performs different gestures in the middle area.
	\item Environment C (tutorial room environment). The furniture is mainly round table and chairs. There is also one big screen on the wall. The two routers are placed on two tables with a distance of 2.2 meters. The volunteer performs different gestures in the testing area.
	\item Environment D (meeting room environment). There is a big table in the center of the room. Chairs are put around the table. We place two routers at one side of the table with a distance of 2.2 meters, and the volunteer performs different gestures beside the table.
\end{enumerate}

We have selected 8 volunteers aging from 19 to 27 to participate in our experiments. 5 of them are males and 3 of them are females. Our experiments involve 8 categories of human gestures including up \& down, left \& right, back \& forward, clap, fist, circling, throw, zoom. We provide the CSI visualization for these gestures in Fig \ref{fig:csiplot}. Each volunteer performs different gestures while the transmitter sends signal packets to the receiver. The CSI enabled platform \cite{yang2018device} measures and stores the CSI data.

For each gesture, 200 CSI samples are recorded at each experimental location. In total, each gesture has 800 CSI samples collected. AirFi only uses CSI amplitude information for gesture recognition. There are 114 subcarriers of each CSI sample received by our router platform during the collection. Each sample consists of 2000 CSI packets, then we downsample the data to 500 packets. The input size of our CSI data is $114\times3\times500$.

\subsection{Overall Evaluation}

We compare the system AirFi with state-of-the-art CSI-based gesture recognition systems. For the compared systems, we select WiGr, WGRDTL and Wi-Multi \cite{zhang2021wifi,bu2018wi,feng2019wi}. The compared systems are selected as they also have the ability to adapt to the new environment by taking the advantage of domain transfer. Besides, we remove the generator components of MCBAR \cite{wang2021multimodal} and only keep the feature extractor and classifier components which are basically convolutional neural networks (CNN). The remaining CNN is able to perform accurate human behavior recognition within one environment. Actually the CNN structure is widely used for classification purpose in many CSI-based human sensing systems. However, it does not have fitting ability to adapt into an unseen environment. We train the CNN to have over 90\% accuracy in the training environment, then we deploy it in the testing environment together with other compared systems. To test the ability of each method in adapting to a new environment, we use the CSI data from three environments as the source environment data and the CSI data from the left environment as the testing environment data. For example, when environment A, B, C are used as training environment, D will be used as the testing environment, and denote the setting as ABC-D. All four different combinations are tested in the experiment. The testing environment data are not available for any systems including AirFi during the training phase.

The experiments results are shown in Table \ref{tab:table1}. As shown in the table, AirFi outperforms all other compared systems in all testing environments with an accuracy of around 90\%. The obvious performance degradation of CNN demonstrates that there are large variations of collected CSI data from different environment settings. Though the compared systems are also designed to adapt to the a environment, they need a large number of testing environment data to perform the domain adaption. As in our scenario, all systems are not able to acquire the testing domain data, their models are not able to be functional as they should be. WiGr has the second best performance with an overall accuracy of over 80\%. In order for it to adapt to a new environment better, it requires CSI data from the target environment. However, this is not provided in the experiment. AirFi takes the advantage of domain generalization. It manages to extract the common feature codes from these source environment data and generalize them on the feature space by minimizing the feature codes distribution differences. The training of AirFi does not need any CSI data from the testing environment.  

\subsection{Impacts of TX-RX Distances}

During the experiments, we have fixed the distances between the two routers from 2.2 meters to 2.3 meters as indicated in the scenario specifications. In fact, the distances between them also affect the system performances. When the distances between the two routers increase, the signal power will decrease and SNR (Signal-Noise Ratio) also becomes lower. This can lead to performance degradation of AirFi.

We examine the impacts of TX-RX distances on performances. We collect CSI data in each scenario with three different sets of distance range:1.8 to 1.9 meters, 2.2 to 2.3 meters, 2.6 to 2.7 meters. Then we use these samples to train AirFi separately and observe the result variations. We found that no obvious performance degradation can be observed between the first two sets. They are both within the functional range of the routers. While for the second and third distance sets, the results of average accuracy of all involved gestures drop about 3\%. When TX-RX distances increase, the performances of AirFi remain stable within the functional range. However, if the distances get further, the performance will degrade.

\subsection{Ablation Study}

In order to equip the system AirFi with the ability to generalize to new environment settings, we take the advantage of domain generalization. We also augment our CSI datasets and feature codes to improve the performance. We use an ablation test to study how each component contributes to the system AirFi. We compare the performances between the pure CNN, AirFi without data augmentation, AirFi without feature augmentation and complete AirFi. The experiments setups are the same as the previous overall evaluation. In each test, three of them are used as the training environments and the remaining one as the testing environment. 

The testing results are shown in Table \ref{tab:table2}. CNN performs worst among the compared systems, while AirFi is still able to generalize its system model which is the most important ability of AirFi. To generalize the system model of AirFi, it needs the CSI data from different training environments. Both CSI data and feature augmentation are used to further enhance the generalization ability. Without these two techniques, the remaining parts of AirFi can still generalize its model and outperform the CNN which does not have any ability to adapt to a new environment. For AirFi without either data augmentation or feature augmentation, their accuracies all get worse compared to the complete AirFi. In other words, both augmentation techniques contribute to the performances of AirFi. It is observed that after removing either data augmentation or feature augmentation, the performance degradation of the two compared systems are very close to each other, which is about 4\% to 6\%. The missing of feature augmentation affects AirFi slightly more than data augmentation. In fact, both augmentation techniques make the feature codes of training to be more representative. With diverse feature codes, AirFi is able to generalize better on the feature space. The data augmentation improves the diversity of CSI training datasets. It generates more simulated CSI data so that more feature codes can be extracted from these generated CSI data. On the other hand, the feature augmentation works directly on the feature codes. It makes the feature codes more representative. With the help of these two augmentation techniques, the feature codes generalized on feature space are more diverse and AirFi has a higher possibility to generalize to a new environment setting.

\begin{figure*}[tb]
	\centering
	\subfigure[CSI features without DG]{
		\begin{minipage}{0.24\linewidth}
			\centering
			\includegraphics[width=4cm, height=3.4cm]{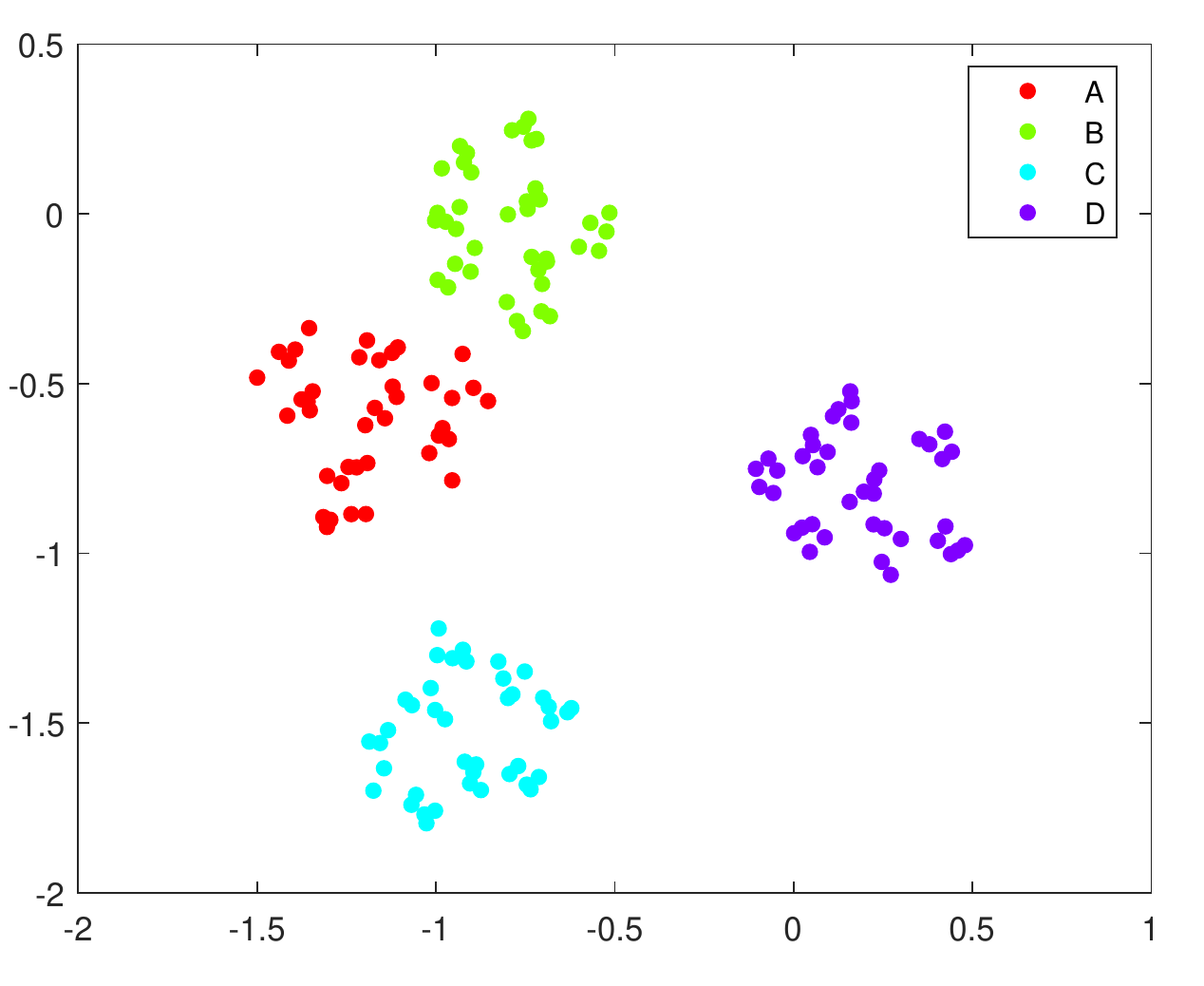}
			\label{fig:tsne1}
		\end{minipage}%
	}%
	\subfigure[CSI features with DG]{
		\begin{minipage}{0.24\linewidth}
			\centering
			\includegraphics[width=4cm, height=3.4cm]{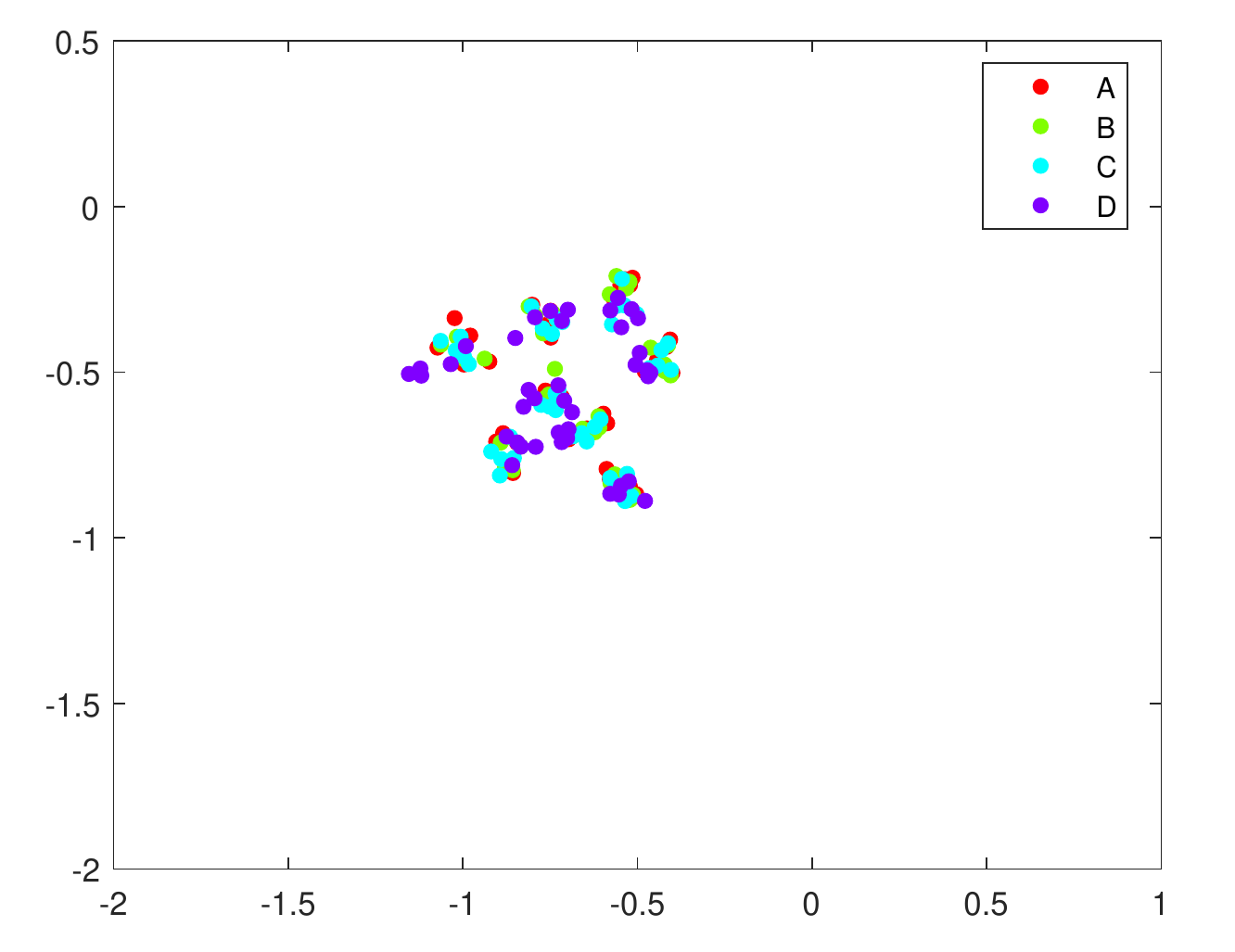}
			\label{fig:tsne2}
		\end{minipage}%
	}%
	\subfigure[CSI features without Data Aug]{
		\begin{minipage}{0.24\linewidth}
			\centering
			\includegraphics[width=4cm, height=3.3cm]{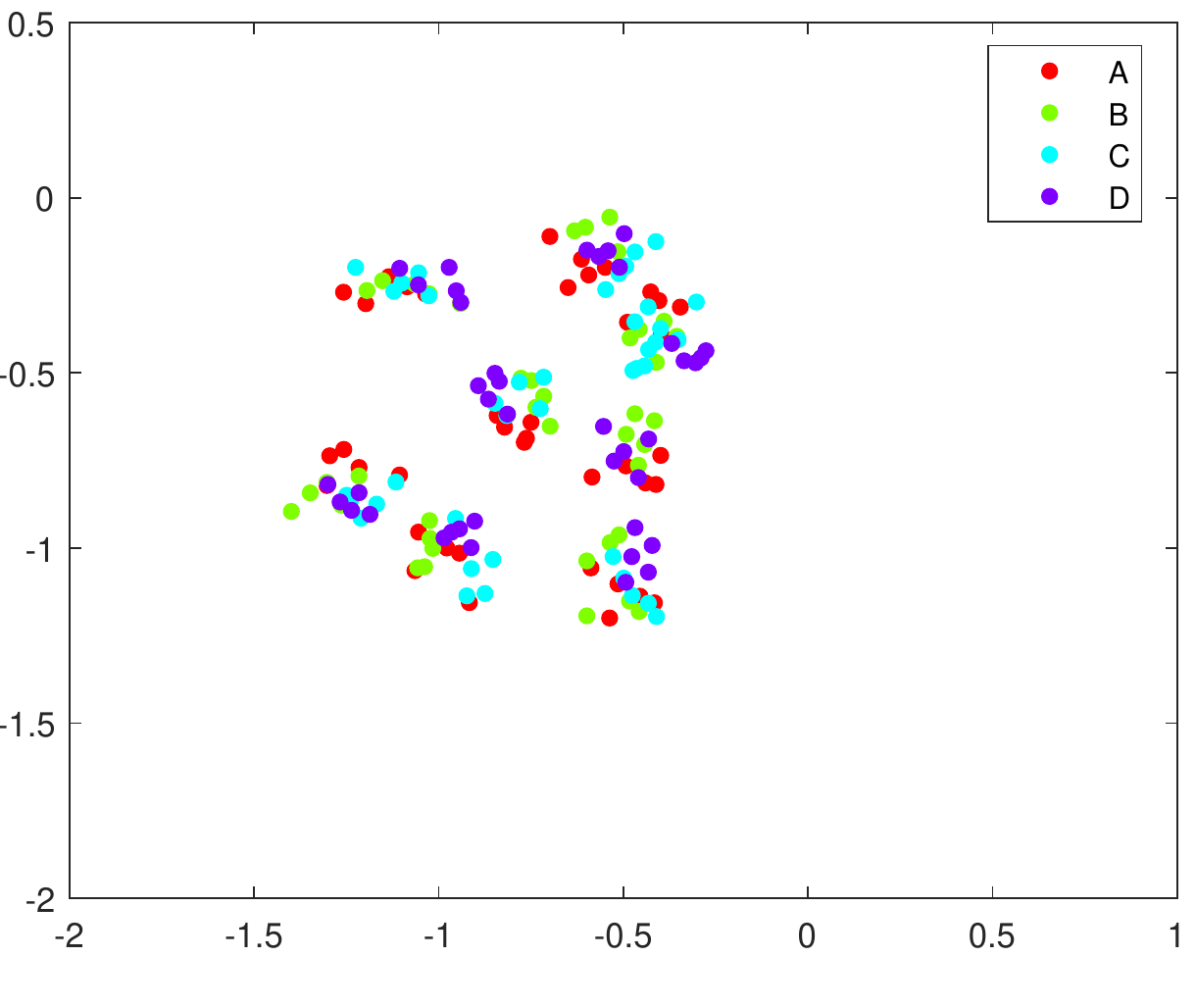}
			\label{fig:tsneData}
		\end{minipage}%
	}%
	\subfigure[CSI features without Feature Aug]{
		\begin{minipage}{0.24\linewidth}
			\centering
			\includegraphics[width=4cm, height=3.4cm]{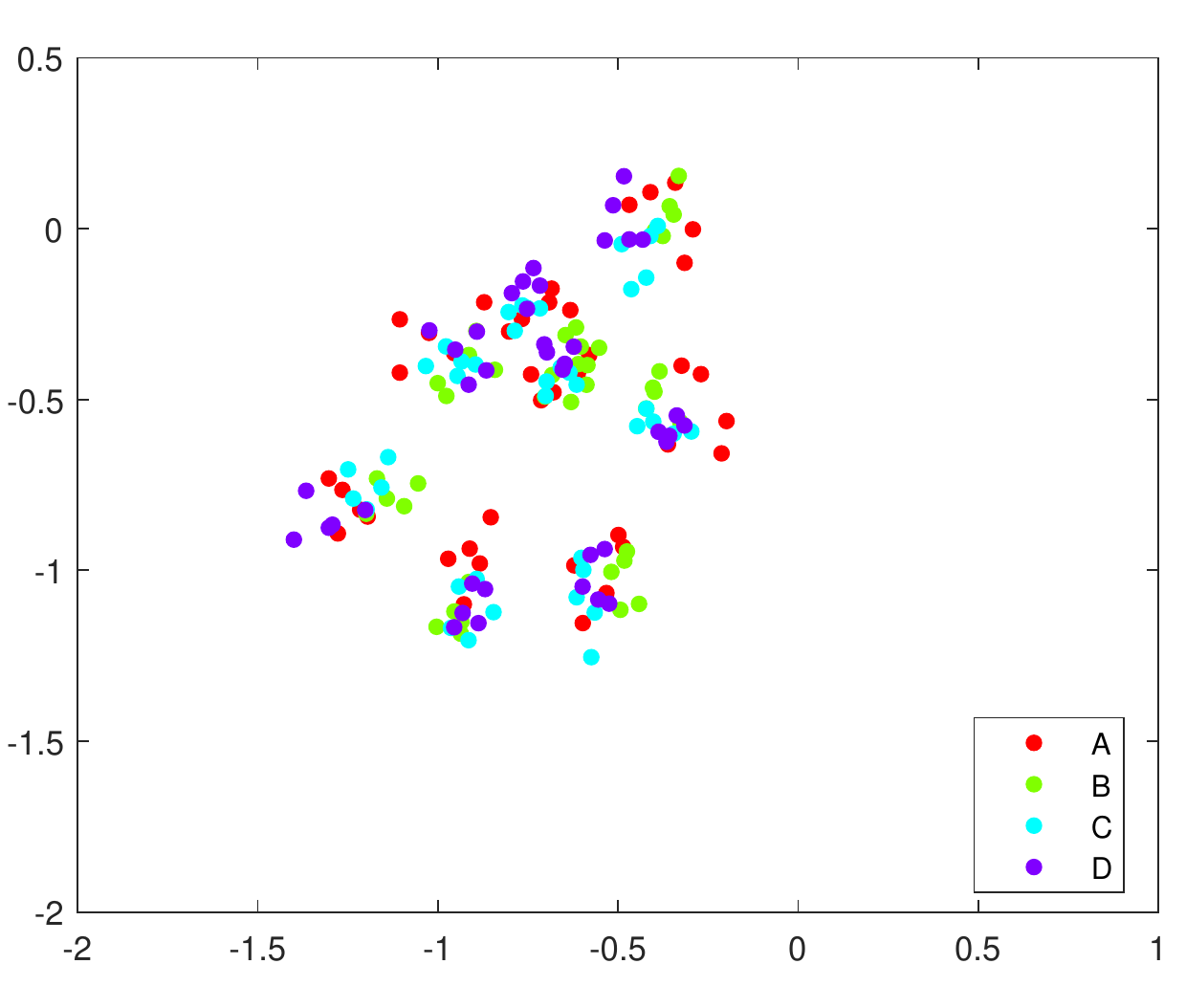}
			\label{fig:tsneFeature}
		\end{minipage}%
	}%
	\centering
	\caption{T-SNE plotting of CSI deep features in different settings.}
	\label{fig:unbandb}
\end{figure*}

\subsection{Few Shot Learning Adds on}

In the previous evaluations, CSI data from the testing domain environments are totally not available during the training phase. We also explore the situation that only a few CSI data from the testing domain environments are used for system training. As for the compared systems, they use domain adaption techniques which require a large number of CSI data to transfer their model to the new environment. We improve AirFi with a few-shot learning technique added on so that the generalized model can be further enhanced using a few labeled CSI samples. For the evaluation, this time, 10 CSI samples from the testing domain environments are available during the training phase for all systems. The environment settings and compared systems are the same as they are in the overall evaluation.

The results are shown in Table \ref{tab:table3}. Obvious improvement of performances can be observed for all systems. AirFi still outperforms the other compared systems. As the amount of the given CSI data is very small, the compared systems do not have enough CSI data to fully retrain their models. As WiGr is also equipped with the few-shot learning property in their prototypical model, it performs the second best among the compared systems. While for AirFi, its system model is already generalized to different environments. By applying the few shot learning techniques, AirFi is able to improve its performances with a limited amount of data and generalize even better to the testing environment. AirFi manages to minimize the distribution difference between the given CSI data from the testing environments and the training environments, which can be achieved with small amounts of CSI data.

\subsection{Distribution Visualization}

To better understand how AirFi can generalize its model to different environments, we use the T-SNE plotting to visualize the distribution of feature codes \cite{van2008visualizing}. We plot the hidden features of CSI data from four environments, which are environment A to D.

As shown in Fig \ref{fig:tsne1}, for a trained system without domain generalization (DG) which can be a convolutional neural network, the hidden features of CSI data from one environment are gathered together while they are away from those of other different environments. When it is applied to a new environment, its model cannot recognize CSI data as the distribution between them is very large. For AirFi which is equipped with the ability of domain generalization, feature codes from different environment settings are gathered together as their distribution differences are minimized during the training phase. This is shown in Fig \ref{fig:tsne2}. As a result, the model trained has a high probability to generalize to the new environment. CSI features visualizations are also provided for the ablation study. Fig \ref{fig:tsneData} illustrates the feature visualization for AirFi without data augmentation and Fig \ref{fig:tsneFeature} is for AirFi without feature augmentation. Both of them achieve worse generalization results than the complete AirFi. Data augmentation affects the AirFi less, while feature augmentation affects the performance more.

\begin{table}[htb]
	\begin{center}
		\caption{Evaluation on Widar 3.0 Dataset}
		\begin{tabular}{c|c|c}
			\toprule
			\textbf{Sytems}  &\textbf{Overall Performances}& \textbf{Environment Index } \\
			\midrule
			{\textbf{WiGr \cite{zhang2021wifi}}} &87.86\% &  \\
			{\textbf{WGRDTL \cite{bu2018wi}}}  &82.64\% & EF-G \\
			{\textbf{Wi-Multi \cite{feng2019wi}}}& 81.31\% &  \\
			{\textbf{AirFi}} & \textbf{95.16\%} &\\
			\midrule
			{\textbf{WiGr \cite{zhang2021wifi}}} &86.32\% &  \\
			{\textbf{WGRDTL \cite{bu2018wi}}}  &80.04\% & EG-F \\
			{\textbf{Wi-Multi \cite{feng2019wi}}}& 82.48\% &  \\
			{\textbf{AirFi}} & \textbf{93.62\%}  &\\
			\midrule
			{\textbf{WiGr \cite{zhang2021wifi}}} &88.62\% &  \\
			{\textbf{WGRDTL \cite{bu2018wi}}}  &   82.44\% & FG-E \\
			{\textbf{Wi-Multi \cite{feng2019wi}}}& 80.12\% &  \\
			{\textbf{AirFi}} & \textbf{94.87\%} &\\
			\bottomrule
			
		\end{tabular}
		\label{tab:table4}
	\end{center}
\end{table}

\subsection{Public Dataset Evaluation}

To further evaluate the performances of AirFi in different environments and CSI platforms, we use the public datasets of Widar 3.0 \cite{zheng2019zero} collected by Intel 5300 devices. It includes CSI data of human gestures from 3 different environments which are an empty classroom (E), a spacious hall (F) and an office room (G). As AirFi needs multiple environments as the source environments, we select two environments as source domains and the remaining environment is used as the target domain for each round of environments. In order to use the Widar 3.0 datasets, we modify the sizes of the convolutional neural network in the feature extractor to fit the data size. Then the hyper-parameters in the model training are kept as same as the experiments on our own datasets, without extra tuning. AirFi uses the feature extractor to extract feature code from the source environment CSI data and use them to train the generalized system model. Then the model is applied in the target environment directly to test its performances. The same set of compared systems in the overall evaluation is used. As shown in Table~\ref{tab:table4}, our AirFi achieves significant improvement against existing methods. It learns the common features from CSI data from the two source environments and builds a generalized gesture recognition model which can be applied to the target environment and different subject in Widar 3.0. This demonstrates that the AirFi can be utilized in multiple WiFi sensing platforms, i.e., Atheros CSI Tool and Intel 5300 NIC, and multiple human sensing tasks, i.e., gesture and activity recognition.

\section{CONCLUSION}
	
This paper has investigated the problem that a CSI-based human sensing system suffers from serious performance degradation under different environments. To deal with this problem, we proposed AirFi which takes the advantage of domain generalization to train a generalized model that can be applied to different environments. Moreover, the training of AirFi does not require CSI data from the testing environment which is more suitable for the real-world situation. The experimental results show that AirFi outperforms the state-of-the-art in this field. For future research direction, we may explore on improve the training efficiency of AirFi. AirFi requires CSI data from multiple source environments to train its generalized model. However, CSI data from different environments may not always be available. We want to reduce the dependency on the number of source environments and train the generalized system model with fewer data domains. Besides, AirFi only explores on using amplitude information for gesture recognition. We will also investigate on how phase information can contribute to the CSI-based smart human sensing.

	\section*{Acknowledgment}
	
	This research of the first author is supported by Agency for Science, Technology and Research (Singapore) under AGS scholarship.

\bibliographystyle{IEEEtran}
\bibliography{Megerliter05,newreference}

\begin{thebibliography}{10}
\providecommand{\url}[1]{#1}
\csname url@samestyle\endcsname
\providecommand{\newblock}{\relax}
\providecommand{\bibinfo}[2]{#2}
\providecommand{\BIBentrySTDinterwordspacing}{\spaceskip=0pt\relax}
\providecommand{\BIBentryALTinterwordstretchfactor}{4}
\providecommand{\BIBentryALTinterwordspacing}{\spaceskip=\fontdimen2\font plus
\BIBentryALTinterwordstretchfactor\fontdimen3\font minus
  \fontdimen4\font\relax}
\providecommand{\BIBforeignlanguage}[2]{{%
\expandafter\ifx\csname l@#1\endcsname\relax
\typeout{** WARNING: IEEEtran.bst: No hyphenation pattern has been}%
\typeout{** loaded for the language `#1'. Using the pattern for}%
\typeout{** the default language instead.}%
\else
\language=\csname l@#1\endcsname
\fi
#2}}
\providecommand{\BIBdecl}{\relax}
\BIBdecl

\bibitem{8941292}
L.~Wang, K.~Sun, H.~Dai, W.~Wang, K.~Huang, A.~X. Liu, X.~Wang, and Q.~Gu,
  ``{WiTrace: Centimeter-Level Passive Gesture Tracking Using OFDM Signals},''
  \emph{{IEEE Transactions on Mobile Computing}}, vol.~20, no.~4, pp.
  1730--1745, 2021.

\bibitem{9115830}
Y.~Chen, R.~Ou, Z.~Li, and K.~Wu, ``{WiFace: Facial Expression Recognition
  Using Wi-Fi Signals},'' \emph{{IEEE Transactions on Mobile Computing}},
  vol.~21, no.~1, pp. 378--391, 2022.

\bibitem{8519328}
H.~Abdelnasser, K.~Harras, and M.~Youssef, ``{A Ubiquitous WiFi-Based
  Fine-Grained Gesture Recognition System},'' \emph{IEEE Transactions on Mobile
  Computing}, vol.~18, no.~11, pp. 2474--2487, 2019.

\bibitem{7345587}
X.~Liu, J.~Cao, S.~Tang, J.~Wen, and P.~Guo, ``{Contactless Respiration
  Monitoring Via Off-the-Shelf WiFi Devices},'' \emph{{IEEE Transactions on
  Mobile Computing}}, vol.~15, no.~10, pp. 2466--2479, 2016.

\bibitem{yang2022deep}
J.~Yang, X.~Chen, D.~Wang, H.~Zou, C.~X. Lu, S.~Sun, and L.~Xie, ``Deep
  learning and its applications to wifi human sensing: A benchmark and a
  tutorial,'' \emph{arXiv preprint arXiv:2207.07859}, 2022.

\bibitem{halperin2010predictable}
D.~Halperin, W.~Hu, A.~Sheth, and D.~Wetherall, ``{Predictable 802.11 Packet
  Delivery from Wireless Channel Measurements},'' \emph{{ACM SIGCOMM Computer
  Communication Review}}, vol.~40, no.~4, pp. 159--170, 2010.

\bibitem{hu2017new}
H.~Hu and L.~Li, ``{A New Method using Covariance Eigenvalues and Time Window
  in Passive Human Motion Detection based on CSI Phases},'' in \emph{{2017 IEEE
  5th International Symposium on Electromagnetic Compatibility
  (EMC-Beijing)}}.\hskip 1em plus 0.5em minus 0.4em\relax IEEE, 2017, pp. 1--6.

\bibitem{halperin2011tool}
D.~Halperin, W.~Hu, A.~Sheth, and D.~Wetherall, ``{Tool Release: Gathering
  802.11 n Traces with Channel State Information},'' \emph{{ACM SIGCOMM
  Computer Communication Review}}, vol.~41, no.~1, pp. 53--53, 2011.

\bibitem{zou2017freecount}
H.~Zou, Y.~Zhou, J.~Yang, W.~Gu, L.~Xie, and C.~Spanos, ``Freecount:
  Device-free crowd counting with commodity wifi,'' in \emph{GLOBECOM 2017-2017
  IEEE Global Communications Conference}.\hskip 1em plus 0.5em minus
  0.4em\relax IEEE, 2017, pp. 1--6.

\bibitem{zou2017freedetector}
------, ``Freedetector: Device-free occupancy detection with commodity wifi,''
  in \emph{2017 IEEE International Conference on Sensing, Communication and
  Networking (SECON Workshops)}.\hskip 1em plus 0.5em minus 0.4em\relax IEEE,
  2017, pp. 1--5.

\bibitem{zou2017multiple}
------, ``Multiple kernel representation learning for wifi-based human activity
  recognition,'' in \emph{2017 16th IEEE International Conference on Machine
  Learning and Applications (ICMLA)}.\hskip 1em plus 0.5em minus 0.4em\relax
  IEEE, 2017, pp. 268--274.

\bibitem{zou2017poster}
------, ``Poster: Wifi-based device-free human activity recognition via
  automatic representation learning,'' in \emph{Proceedings of the 23rd annual
  international conference on mobile computing and networking}, 2017, pp.
  606--608.

\bibitem{zou2019wifi}
H.~Zou, J.~Yang, H.~Prasanna~Das, H.~Liu, Y.~Zhou, and C.~J. Spanos, ``Wifi and
  vision multimodal learning for accurate and robust device-free human activity
  recognition,'' in \emph{Proceedings of the IEEE/CVF conference on computer
  vision and pattern recognition workshops}, 2019, pp. 0--0.

\bibitem{zou2018wifi}
H.~Zou, Y.~Zhou, J.~Yang, W.~Gu, L.~Xie, and C.~J. Spanos, ``Wifi-based human
  identification via convex tensor shapelet learning,'' in \emph{Thirty-Second
  AAAI Conference on Artificial Intelligence}, 2018.

\bibitem{zou2018robust}
H.~Zou, J.~Yang, Y.~Zhou, L.~Xie, and C.~J. Spanos, ``Robust wifi-enabled
  device-free gesture recognition via unsupervised adversarial domain
  adaptation,'' in \emph{2018 27th International Conference on Computer
  Communication and Networks (ICCCN)}.\hskip 1em plus 0.5em minus 0.4em\relax
  IEEE, 2018, pp. 1--8.

\bibitem{zou2018joint}
H.~Zou, J.~Yang, Y.~Zhou, and C.~J. Spanos, ``Joint adversarial domain
  adaptation for resilient wifi-enabled device-free gesture recognition,'' in
  \emph{2018 17th IEEE International Conference on Machine Learning and
  Applications (ICMLA)}.\hskip 1em plus 0.5em minus 0.4em\relax IEEE, 2018, pp.
  202--207.

\bibitem{zhang2017toward}
D.~Zhang, H.~Wang, and D.~Wu, ``{Toward Centimeter-scale Human Activity Sensing
  with WiFi Signals},'' \emph{{Computer}}, vol.~50, no.~1, pp. 48--57, 2017.

\bibitem{wang2016human}
H.~Wang, D.~Zhang, J.~Ma, Y.~Wang, Y.~Wang, D.~Wu, T.~Gu, and B.~Xie, ``{Human
  Respiration Detection with Commodity WiFi Devices: Do user location and body
  orientation matter?}'' in \emph{{Proceedings of the 2016 ACM International
  Joint Conference on Pervasive and Ubiquitous Computing}}, 2016, pp. 25--36.

\bibitem{wang2015understanding}
W.~Wang, A.~X. Liu, M.~Shahzad, K.~Ling, and S.~Lu, ``{Understanding and
  Modeling of WiFi Signal based Human Activity Recognition},'' in
  \emph{{Proceedings of the 21st Annual International Conference on Mobile
  Computing and Networking}}, 2015, pp. 65--76.

\bibitem{pu2013whole}
Q.~Pu, S.~Gupta, S.~Gollakota, and S.~Patel, ``{Whole-home Gesture Recognition
  using Wireless Signals},'' in \emph{{Proceedings of the 19th Annual
  International Conference on Mobile Computing and Networking}}, 2013, pp.
  27--38.

\bibitem{daume2009frustratingly}
H.~Daum{\'e}~III, ``{Frustratingly Easy Domain Adaptation},'' \emph{{arXiv
  preprint arXiv:0907.1815}, year={2009}}.

\bibitem{wang2018deep}
M.~Wang and W.~Deng, ``{Deep Visual Domain Adaptation: A Survey},''
  \emph{{Neurocomputing}}, vol. 312, pp. 135--153, 2018.

\bibitem{pan2010domain}
S.~J. Pan, I.~W. Tsang, J.~T. Kwok, and Q.~Yang, ``{Domain Adaptation via
  Transfer Component Analysis},'' \emph{{IEEE Transactions on Neural
  Networks}}, vol.~22, no.~2, pp. 199--210, 2010.

\bibitem{zou2019consensus}
H.~Zou, Y.~Zhou, J.~Yang, H.~Liu, H.~P. Das, and C.~J. Spanos, ``Consensus
  adversarial domain adaptation,'' in \emph{The Thirty-Third AAAI Conference on
  Artificial Intelligence (AAAI-19)}, 2019.

\bibitem{yang2020mind}
J.~Yang, H.~Zou, Y.~Zhou, Z.~Zeng, and L.~Xie, ``Mind the discriminability:
  Asymmetric adversarial domain adaptation,'' in \emph{European Conference on
  Computer Vision}, 2020.

\bibitem{yang2020mobileda}
J.~Yang, H.~Zou, S.~Cao, Z.~Chen, and L.~Xie, ``Mobileda: Toward edge domain
  adaptation,'' \emph{IEEE Internet of Things Journal}, vol.~7, no.~8, pp.
  6909--6918, 2020.

\bibitem{wang2021multimodal}
D.~Wang, J.~Yang, W.~Cui, L.~Xie, and S.~Sun, ``{Multimodal CSI-based Human
  Activity Recognition using GANs},'' \emph{{IEEE Internet of Things Journal}},
  vol.~8, no.~24, pp. 17\,345--17\,355, 2021.

\bibitem{muandet2013domain}
K.~Muandet, D.~Balduzzi, and B.~Sch{\"o}lkopf, ``{Domain generalization via
  Invariant Feature Representation},'' in \emph{{International Conference on
  Machine Learning}}.\hskip 1em plus 0.5em minus 0.4em\relax PMLR, 2013, pp.
  10--18.

\bibitem{li2018learning}
D.~Li, Y.~Yang, Y.-Z. Song, and T.~M. Hospedales, ``{Learning to Generalize:
  Meta-learning for Domain Generalization},'' in \emph{{Thirty-Second AAAI
  Conference on Artificial Intelligence}}, 2018.

\bibitem{zhou2021domain}
K.~Zhou, Y.~Yang, Y.~Qiao, and T.~Xiang, ``{Domain Generalization with
  Mixstyle},'' \emph{arXiv preprint arXiv:2104.02008}, 2021.

\bibitem{li2021two}
B.~Li, W.~Cui, W.~Wang, L.~Zhang, Z.~Chen, and M.~Wu, ``{Two-stream Convolution
  Augmented Transformer for Human Activity Recognition},'' in
  \emph{{Proceedings of the AAAI Conference on Artificial Intelligence}},
  vol.~35, no.~1, 2021, pp. 286--293.

\bibitem{zhang2021privacy}
L.~Zhang, W.~Cui, B.~Li, Z.~Chen, M.~Wu, and T.~S. Gee, ``{Privacy-Preserving
  Cross-Environment Human Activity Recognition},'' \emph{{IEEE Transactions on
  Cybernetics}}, 2021.

\bibitem{9305332}
D.~Wang, J.~Yang, W.~Cui, L.~Xie, and S.~Sun, ``{Robust CSI-based Human
  Activity Recognition using Roaming Generator},'' in \emph{{2020 16th
  International Conference on Control, Automation, Robotics and Vision
  (ICARCV)}}, 2020, pp. 1329--1334.

\bibitem{yang2018fine}
J.~Yang, H.~Zou, H.~Jiang, and L.~Xie, ``Fine-grained adaptive
  location-independent activity recognition using commodity wifi,'' in
  \emph{2018 IEEE Wireless Communications and Networking Conference
  (WCNC)}.\hskip 1em plus 0.5em minus 0.4em\relax IEEE, 2018, pp. 1--6.

\bibitem{yang2018carefi}
------, ``Carefi: Sedentary behavior monitoring system via commodity wifi
  infrastructures,'' \emph{IEEE Transactions on Vehicular Technology}, vol.~67,
  no.~8, pp. 7620--7629, 2018.

\bibitem{deng2022gaitfi}
L.~Deng, J.~Yang, S.~Yuan, H.~Zou, C.~X. Lu, and L.~Xie, ``Gaitfi: Robust
  device-free human identification via wifi and vision multimodal learning,''
  \emph{IEEE Internet of Things Journal}, 2022.

\bibitem{yang2022autofi}
J.~Yang, X.~Chen, H.~Zou, D.~Wang, and L.~Xie, ``Autofi: Towards automatic wifi
  human sensing via geometric self-supervised learning,'' \emph{arXiv preprint
  arXiv:2205.01629}, 2022.

\bibitem{yang2022securesense}
J.~Yang, H.~Zou, and L.~Xie, ``Securesense: Defending adversarial attack for
  secure device-free human activity recognition,'' \emph{IEEE Transactions on
  Mobile Computing}, 2022.

\bibitem{yang2022efficientfi}
J.~Yang, X.~Chen, H.~Zou, D.~Wang, Q.~Xu, and L.~Xie, ``Efficientfi: Towards
  large-scale lightweight wifi sensing via csi compression,'' \emph{IEEE
  Internet of Things Journal}, 2022.

\bibitem{xiong2015csi}
H.~Xiong, F.~Gong, L.~Qu, C.~Du, and K.~Harfoush, ``{CSI-based Device-free
  Gesture Detection},'' in \emph{{2015 12th International Conference on
  High-capacity Optical Networks and Enabling/Emerging Technologies
  (HONET)}}.\hskip 1em plus 0.5em minus 0.4em\relax IEEE, 2015, pp. 1--5.

\bibitem{tan2016wifinger}
S.~Tan and J.~Yang, ``{WiFinger: Leveraging Commodity WiFi for Fine-grained
  Finger Gesture Recognition},'' in \emph{{Proceedings of the 17th ACM
  International Symposium on Mobile Ad hoc Networking and Computing}}, 2016,
  pp. 201--210.

\bibitem{8514811}
Z.~Chen, L.~Zhang, C.~Jiang, Z.~Cao, and W.~Cui, ``{WiFi CSI Based Passive
  Human Activity Recognition Using Attention Based BLSTM},'' \emph{{IEEE
  Transactions on Mobile Computing}}, vol.~18, no.~11, pp. 2714--2724, 2019.

\bibitem{8832182}
T.~Zhang, T.~Song, D.~Chen, T.~Zhang, and J.~Zhuang, ``{WiGrus: A Wifi-Based
  Gesture Recognition System Using Software-Defined Radio},'' \emph{{IEEE
  Access}}, vol.~7, pp. 131\,102--131\,113, 2019.

\bibitem{yang2019learning}
J.~Yang, H.~Zou, Y.~Zhou, and L.~Xie, ``{Learning Gestures from WiFi: A Siamese
  Recurrent Convolutional Architecture},'' \emph{{IEEE Internet of Things
  Journal}}, vol.~6, no.~6, pp. 10\,763--10\,772, 2019.

\bibitem{8726822}
J.~Huang, B.~Liu, H.~Jin, and Z.~Liu, ``{WiAnti: an Anti-Interference Activity
  Recognition System Based on WiFi CSI},'' in \emph{{2018 IEEE International
  Conference on Internet of Things (iThings) and IEEE Green Computing and
  Communications (GreenCom) and IEEE Cyber, Physical and Social Computing
  (CPSCom) and IEEE Smart Data (SmartData)}}, 2018, pp. 58--65.

\bibitem{li2020location}
Y.~Li, T.~Jiang, X.~Ding, and Y.~Wang, ``{Location-free CSI based Activity
  Recognition with Angle Difference of Arrival},'' in \emph{{2020 IEEE Wireless
  Communications and Networking Conference (WCNC)}}.\hskip 1em plus 0.5em minus
  0.4em\relax IEEE, 2020, pp. 1--6.

\bibitem{zheng2019zero}
Y.~Zheng, Y.~Zhang, K.~Qian, G.~Zhang, Y.~Liu, C.~Wu, and Z.~Yang,
  ``Zero-effort cross-domain gesture recognition with wi-fi,'' in
  \emph{Proceedings of the 17th Annual International Conference on Mobile
  Systems, Applications, and Services}, 2019, pp. 313--325.

\bibitem{9141400}
C.~Li, M.~Liu, and Z.~Cao, ``Wihf: Gesture and user recognition with wifi,''
  \emph{IEEE Transactions on Mobile Computing}, vol.~21, no.~2, pp. 757--768,
  2022.

\bibitem{zou2018towards}
H.~Zou, Y.~Zhou, J.~Yang, and C.~J. Spanos, ``Towards occupant activity driven
  smart buildings via wifi-enabled iot devices and deep learning,''
  \emph{Energy and Buildings}, vol. 177, pp. 12--22, 2018.

\bibitem{9145101}
Z.~Shi, J.~A. Zhang, R.~Y. Xu, and Q.~Cheng, ``{WiFi-Based Activity Recognition
  using Activity Filter and Enhanced Correlation with Deep Learning},'' in
  \emph{{2020 IEEE International Conference on Communications Workshops (ICC
  Workshops)}}, 2020, pp. 1--6.

\bibitem{9322627}
Z.~Shi, J.~A. Zhang, R.~Xu, Q.~Cheng, and A.~Pearce, ``{Towards
  Environment-Independent Human Activity Recognition using Deep Learning and
  Enhanced CSI},'' in \emph{{GLOBECOM 2020 - 2020 IEEE Global Communications
  Conference}}, 2020, pp. 1--6.

\bibitem{xiao2019csigan}
C.~Xiao, D.~Han, Y.~Ma, and Z.~Qin, ``{CsiGAN: Robust Channel State
  Information-based Activity Recognition with GANs},'' \emph{{IEEE Internet of
  Things Journal}}, vol.~6, no.~6, pp. 10\,191--10\,204, 2019.

\bibitem{zhang2018crosssense}
J.~Zhang, Z.~Tang, M.~Li, D.~Fang, P.~Nurmi, and Z.~Wang, ``{CrossSense:
  Towards Cross-site and Large-scale WiFi sensing},'' in \emph{{Proceedings of
  the 24th Annual International Conference on Mobile Computing and
  Networking}}, 2018, pp. 305--320.

\bibitem{jiang2018towards}
W.~Jiang, C.~Miao, F.~Ma, S.~Yao, Y.~Wang, Y.~Yuan, H.~Xue, C.~Song, X.~Ma,
  D.~Koutsonikolas \emph{et~al.}, ``Towards environment independent device free
  human activity recognition,'' in \emph{Proceedings of the 24th annual
  international conference on mobile computing and networking}, 2018, pp.
  289--304.

\bibitem{tzeng2017adversarial}
E.~Tzeng, J.~Hoffman, K.~Saenko, and T.~Darrell, ``{Adversarial Discriminative
  Domain Adaptation},'' in \emph{{Proceedings of the IEEE Conference on
  Computer Vision and Pattern Recognition}}, 2017, pp. 7167--7176.

\bibitem{tzeng2014deep}
E.~Tzeng, J.~Hoffman, N.~Zhang, K.~Saenko, and T.~Darrell, ``{Deep Domain
  Confusion: Maximizing for Domain Invariance},'' \emph{arXiv preprint
  arXiv:1412.3474}, 2014.

\bibitem{long2015learning}
M.~Long, Y.~Cao, J.~Wang, and M.~Jordan, ``{Learning Transferable Features with
  Deep Adaptation Networks},'' in \emph{{International Conference on Machine
  Learning}}.\hskip 1em plus 0.5em minus 0.4em\relax PMLR, 2015, pp. 97--105.

\bibitem{liu2017unsupervised}
M.-Y. Liu, T.~Breuel, and J.~Kautz, ``{Unsupervised Image-to-Image Translation
  Networks},'' \emph{{Advances in Neural Information Processing Systems}},
  vol.~30, 2017.

\bibitem{huang2018multimodal}
X.~Huang, M.-Y. Liu, S.~Belongie, and J.~Kautz, ``{Multimodal Unsupervised
  Image-to-Image Translation},'' in \emph{{Proceedings of the European
  Conference on Computer Vision (ECCV)}}, 2018, pp. 172--189.

\bibitem{liu2016coupled}
M.-Y. Liu and O.~Tuzel, ``{Coupled Generative Adversarial Networks},''
  \emph{{Advances in Neural Information Processing Systems}}, vol.~29, 2016.

\bibitem{li2018domain}
H.~Li, S.~J. Pan, S.~Wang, and A.~C. Kot, ``{Domain Generalization with
  Adversarial Feature Learning},'' in \emph{{Proceedings of the IEEE Conference
  on Computer Vision and Pattern Recognition}}, 2018, pp. 5400--5409.

\bibitem{blanchard2011generalizing}
G.~Blanchard, G.~Lee, and C.~Scott, ``{Generalizing from Several Related
  Classification Tasks to A New Unlabeled Sample},'' \emph{{Advances in Neural
  Information Processing Systems}}, vol.~24, 2011.

\bibitem{khosla2012undoing}
A.~Khosla, T.~Zhou, T.~Malisiewicz, A.~A. Efros, and A.~Torralba, ``{Undoing
  the Damage of Dataset Bias},'' in \emph{{European Conference on Computer
  Vision}}.\hskip 1em plus 0.5em minus 0.4em\relax Springer, 2012, pp.
  158--171.

\bibitem{goodfellow2014generative}
I.~Goodfellow, J.~Pouget-Abadie, M.~Mirza, B.~Xu, D.~Warde-Farley, S.~Ozair,
  A.~Courville, and Y.~Bengio, ``{Generative Adversarial Nets},''
  \emph{{Advances in Neural Information Processing Systems}}, vol.~27, 2014.

\bibitem{li2015generative}
Y.~Li, K.~Swersky, and R.~Zemel, ``{Generative Moment Matching Networks},'' in
  \emph{{International Conference on Machine Learning}}.\hskip 1em plus 0.5em
  minus 0.4em\relax PMLR, 2015, pp. 1718--1727.

\bibitem{li2018deep}
Y.~Li, X.~Tian, M.~Gong, Y.~Liu, T.~Liu, K.~Zhang, and D.~Tao, ``{Deep Domain
  Generalization via Conditional Invariant Adversarial Networks},'' in
  \emph{{Proceedings of the European Conference on Computer Vision (ECCV)}},
  2018, pp. 624--639.

\bibitem{wang2020generalizing}
Y.~Wang, Q.~Yao, J.~T. Kwok, and L.~M. Ni, ``{Generalizing from A Few Examples:
  A survey on Few-shot Learning},'' \emph{{ACM Computing Surveys (csur)}},
  vol.~53, no.~3, pp. 1--34, 2020.

\bibitem{goldberger2004neighbourhood}
J.~Goldberger, G.~E. Hinton, S.~Roweis, and R.~R. Salakhutdinov,
  ``{Neighbourhood Components Analysis},'' \emph{{Advances in Neural
  Information Processing Systems}}, vol.~17, pp. 513--520, 2004.

\bibitem{salakhutdinov2007learning}
R.~Salakhutdinov and G.~Hinton, ``{Learning a Nonlinear Embedding by Preserving
  Class Neighbourhood Structure},'' in \emph{{Artificial Intelligence and
  Statistics}}.\hskip 1em plus 0.5em minus 0.4em\relax PMLR, 2007, pp.
  412--419.

\bibitem{weinberger2009distance}
K.~Q. Weinberger and L.~K. Saul, ``{Distance Metric Learning for Large Margin
  Nearest Neighbor Classification},'' \emph{{Journal of Machine Learning
  Research}}, vol.~10, no.~2, 2009.

\bibitem{min2009deep}
R.~Min, D.~A. Stanley, Z.~Yuan, A.~Bonner, and Z.~Zhang, ``{A Deep Non-linear
  Feature Mapping for Large-margin KNN Classification},'' in \emph{{2009 Ninth
  IEEE International Conference on Data Mining}}.\hskip 1em plus 0.5em minus
  0.4em\relax IEEE, 2009, pp. 357--366.

\bibitem{snell2017prototypical}
J.~Snell, K.~Swersky, and R.~S. Zemel, ``{Prototypical Networks for Few-shot
  Learning},'' \emph{arXiv preprint arXiv:1703.05175}, 2017.

\bibitem{koch2015siamese}
G.~Koch, R.~Zemel, R.~Salakhutdinov \emph{et~al.}, ``{Siamese Neural Networks
  for One-shot Image Recognition},'' in \emph{{ICML Deep Learning Workshop}},
  vol.~2.\hskip 1em plus 0.5em minus 0.4em\relax Lille, 2015, p.~0.

\bibitem{sung2018learning}
F.~Sung, Y.~Yang, L.~Zhang, T.~Xiang, P.~H. Torr, and T.~M. Hospedales,
  ``{Learning to Compare: Relation Network for Few-shot Learning},'' in
  \emph{{Proceedings of the IEEE Conference on Computer Vision and Pattern
  Recognition}}, 2018, pp. 1199--1208.

\bibitem{yosinski2014transferable}
J.~Yosinski, J.~Clune, Y.~Bengio, and H.~Lipson, ``{How Transferable Are
  Features in Deep Neural Networks?}'' \emph{{Advances in Neural Information
  Processing Systems}}, vol.~27, 2014.

\bibitem{bateni2020improved}
P.~Bateni, R.~Goyal, V.~Masrani, F.~Wood, and L.~Sigal, ``{Improved Few-shot
  Visual Classification},'' in \emph{{Proceedings of the IEEE/CVF Conference on
  Computer Vision and Pattern Recognition}}, 2020, pp. 14\,493--14\,502.

\bibitem{makhzani2015adversarial}
A.~Makhzani, J.~Shlens, N.~Jaitly, I.~Goodfellow, and B.~Frey, ``{Adversarial
  Autoencoders},'' \emph{arXiv preprint arXiv:1511.05644}, 2015.

\bibitem{zheng2016improving}
S.~Zheng, Y.~Song, T.~Leung, and I.~Goodfellow, ``{Improving the Robustness of
  Deep Neural Networks via Stability Training},'' in \emph{{Proceedings of the
  IEEE Conference on Computer Vision and Pattern Recognition}}, 2016, pp.
  4480--4488.

\bibitem{li2021simple}
P.~Li, D.~Li, W.~Li, S.~Gong, Y.~Fu, and T.~M. Hospedales, ``{A Simple Feature
  Augmentation for Domain Generalization},'' in \emph{{Proceedings of the
  IEEE/CVF International Conference on Computer Vision}}, 2021, pp. 8886--8895.

\bibitem{li2021metasaug}
S.~Li, K.~Gong, C.~H. Liu, Y.~Wang, F.~Qiao, and X.~Cheng, ``Metasaug: Meta
  semantic augmentation for long-tailed visual recognition,'' in
  \emph{Proceedings of the IEEE/CVF conference on computer vision and pattern
  recognition}, 2021, pp. 5212--5221.

\bibitem{li2021transferable}
S.~Li, M.~Xie, K.~Gong, C.~H. Liu, Y.~Wang, and W.~Li, ``Transferable semantic
  augmentation for domain adaptation,'' in \emph{Proceedings of the IEEE/CVF
  Conference on Computer Vision and Pattern Recognition}, 2021, pp.
  11\,516--11\,525.

\bibitem{wang2022generalizing}
J.~Wang, C.~Lan, C.~Liu, Y.~Ouyang, T.~Qin, W.~Lu, Y.~Chen, W.~Zeng, and P.~Yu,
  ``Generalizing to unseen domains: A survey on domain generalization,''
  \emph{IEEE Transactions on Knowledge and Data Engineering}, 2022.

\bibitem{smola2007hilbert}
A.~Smola, A.~Gretton, L.~Song, and B.~Sch{\"o}lkopf, ``{A Hilbert Space
  Embedding for Distributions},'' in \emph{{International Conference on
  Algorithmic Learning Theory}}.\hskip 1em plus 0.5em minus 0.4em\relax
  Springer, 2007, pp. 13--31.

\bibitem{zhang2021wifi}
X.~Zhang, C.~Tang, K.~Yin, and Q.~Ni, ``{WiFi-based Cross-Domain Gesture
  Recognition via Modified Prototypical Networks},'' \emph{{IEEE Internet of
  Things Journal}}, 2021.

\bibitem{bu2018wi}
Q.~Bu, G.~Yang, J.~Feng, and X.~Ming, ``{Wi-fi based Gesture Recognition using
  Deep Transfer Learning},'' in \emph{2018 IEEE SmartWorld, Ubiquitous
  Intelligence \& Computing, Advanced \& Trusted Computing, Scalable Computing
  \& Communications, Cloud \& Big Data Computing, Internet of People and Smart
  City Innovation (SmartWorld/SCALCOM/UIC/ATC/CBDCom/IOP/SCI)}.\hskip 1em plus
  0.5em minus 0.4em\relax IEEE, 2018, pp. 590--595.

\bibitem{feng2019wi}
C.~Feng, S.~Arshad, S.~Zhou, D.~Cao, and Y.~Liu, ``{Wi-multi: A Three-phase
  System for Multiple Human Activity Recognition with Commercial WiFi
  Devices},'' \emph{{IEEE Internet of Things Journal}}, vol.~6, no.~4, pp.
  7293--7304, 2019.

\bibitem{yang2018device}
J.~Yang, H.~Zou, H.~Jiang, and L.~Xie, ``{Device-free Occupant Activity Sensing
  using WiFi-enabled IoT Devices for Smart Homes},'' \emph{{IEEE Internet of
  Things Journal}}, vol.~5, no.~5, pp. 3991--4002, 2018.

\bibitem{van2008visualizing}
L.~Van~der Maaten and G.~Hinton, ``{Visualizing Data using t-SNE},''
  \emph{{Journal of Machine Learning Research}}, vol.~9, no.~11, 2008.

\end{thebibliography}

\begin{IEEEbiography}[{\includegraphics[width=1in,height=1.25in,clip,keepaspectratio]{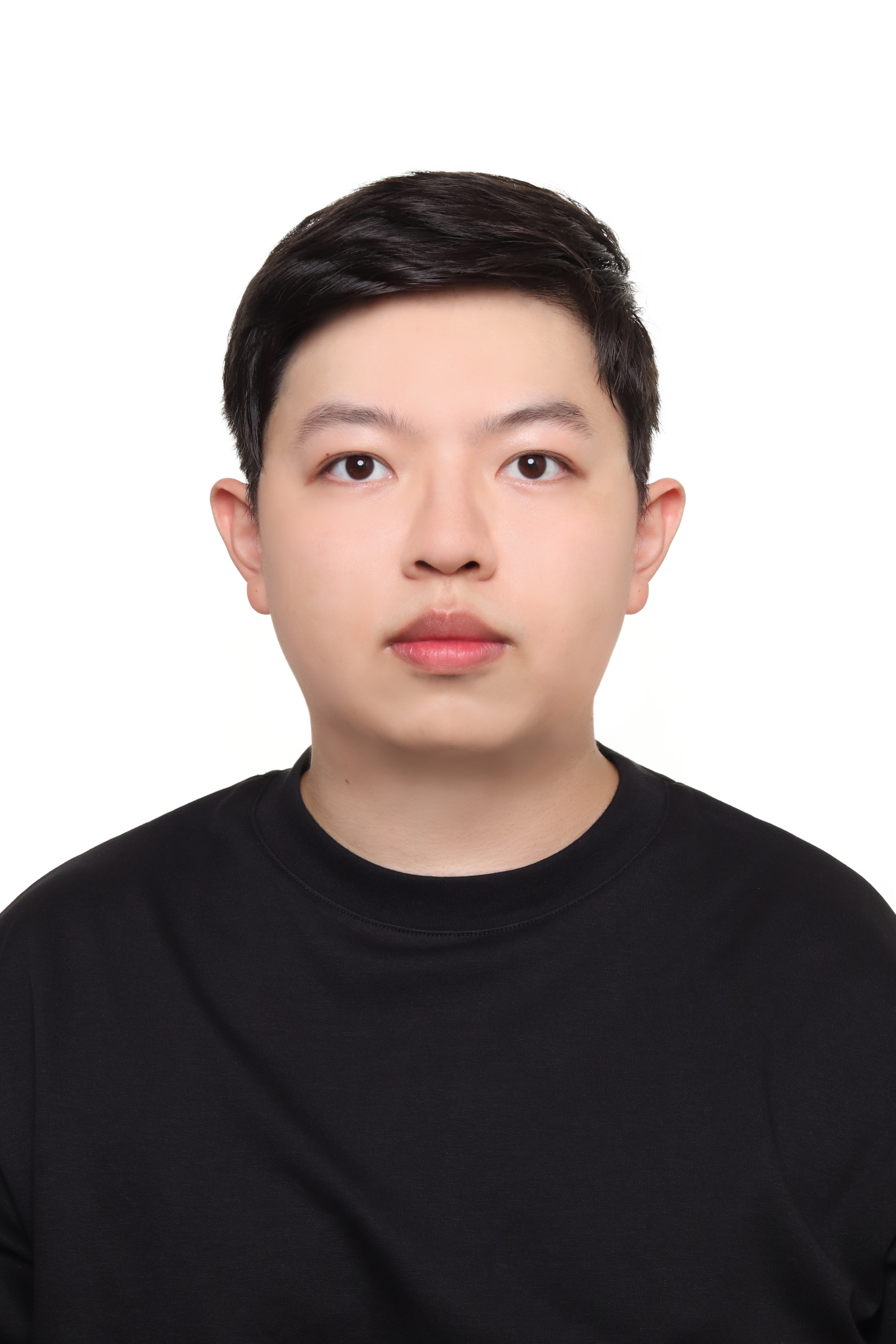}}]{Dazhuo Wang}
	received the B.Eng. from the School of Electrical and Electronic Engineering, Nanyang Technological University in 2018. He is currently a PhD candidate in the School of Electrical and Electronic Engineering, Nanyang Technological University, Singapore. Besides, he is also currently a scientist at Institute for Infocomm Research, Agency for Science, Technology and Research (A*STAR), Singapore. His research interests include Industrial Internet of Things and machine learning. 
\end{IEEEbiography}
\vskip 0pt plus -1fil 
\begin{IEEEbiography}[{\includegraphics[width=0.9in,clip,keepaspectratio]{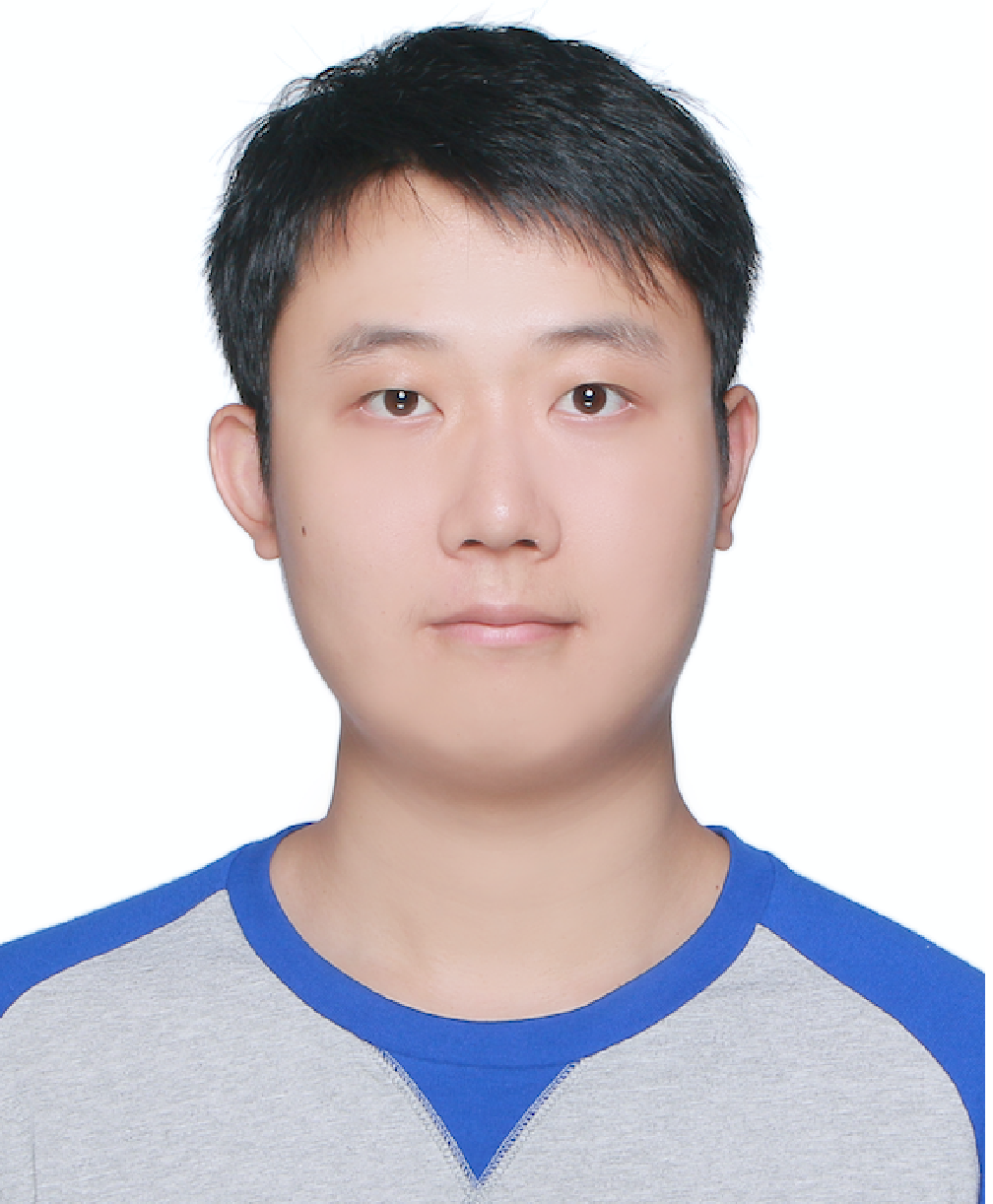}}]{Jianfei Yang}
	received the B.Eng. from the School of Data and Computer Science, Sun Yat-sen University in 2016. He is currently a PhD candidate in the School of Electrical and Electronic Engineering, Nanyang Technological University, Singapore. His research interests include deep transfer learning with applications in Internet of Things and computer vision. He won many AI and data challenges in the visual and interdisciplinary fields, such as ACM ICMI EmotiW-18 and IEEE CVPR-19 UG2+ challenge.
\end{IEEEbiography}
\vskip 0pt plus -1fil 
\begin{IEEEbiography}[{\includegraphics[width=1in,height=1.25in,clip,keepaspectratio]{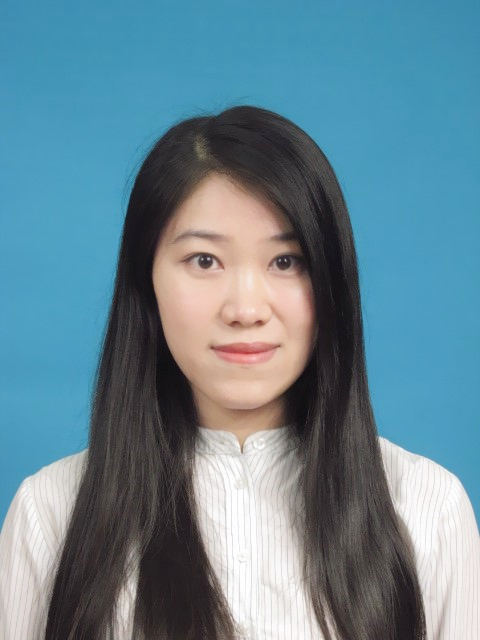}}]{Wei Cui} received the M.E. and Ph.D. degrees in pattern recognition and intelligent system from Northeastern University, Shenyang, China, in 2013 and 2017, respectively. Currently, she is a scientist at Institute for Infocomm Research, Agency for Science, Technology and Research (A*STAR), Singapore. Her current research interests include wireless sensor networks, localization and navigation, and machine learning.
\end{IEEEbiography}

\vskip 0pt plus -1fil 
\begin{IEEEbiography}[{\includegraphics[width=1in,height=1.25in,clip,keepaspectratio]{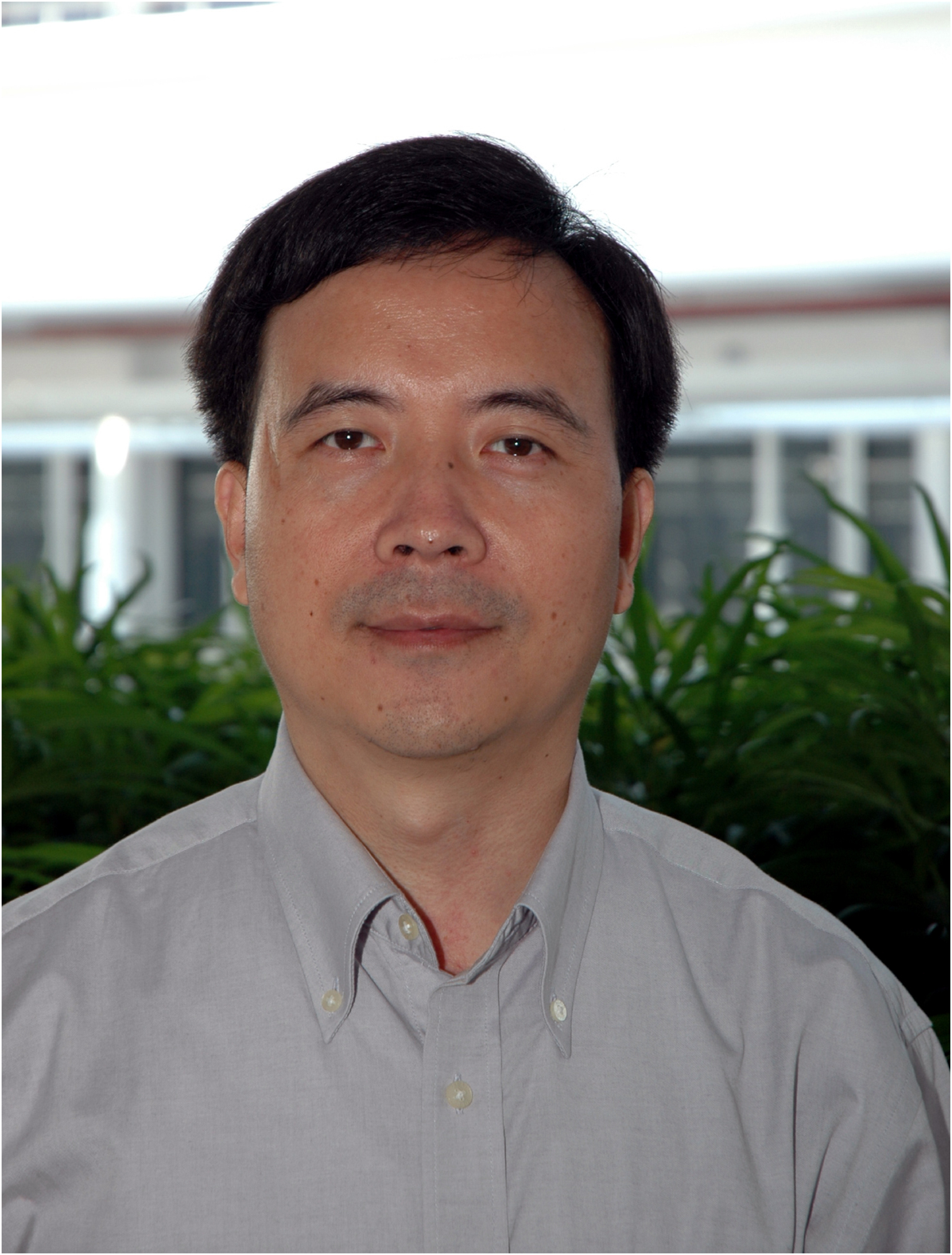}}]{Lihua Xie}
	received the B.E. and M.E. degrees in electrical engineering from Nanjing University of Science and Technology in 1983 and 1986, respectively, and the Ph.D. degree in electrical engineering from the University of Newcastle, Australia, in 1992. Since 1992, he has been with the School of Electrical and Electronic Engineering, Nanyang Technological University, Singapore, where he is currently a professor and served as the Head of Division of Control and Instrumentation from July 2011 to June 2014. He held teaching appointments in the Department of Automatic Control, Nanjing University of Science and Technology from 1986 to 1989 and Changjiang Visiting Professorship with South China University of  Technology from 2006 to 2011.
	
	Dr Xie's research interests include robust control and estimation, networked control systems, multi-agent control and unmanned systems. He has served as an editor of IET Book Series in Control and an Associate Editor of a number of journals including IEEE Transactions on Automatic Control, Automatica, IEEE Transactions on Control Systems Technology, and IEEE Transactions on Circuits and Systems-II. Dr Xie is a Fellow of IEEE and Fellow of IFAC.
\end{IEEEbiography}
\vskip 0pt plus -1fil 
\begin{IEEEbiography}[{\includegraphics[width=1in,height=1.25in,clip,keepaspectratio]{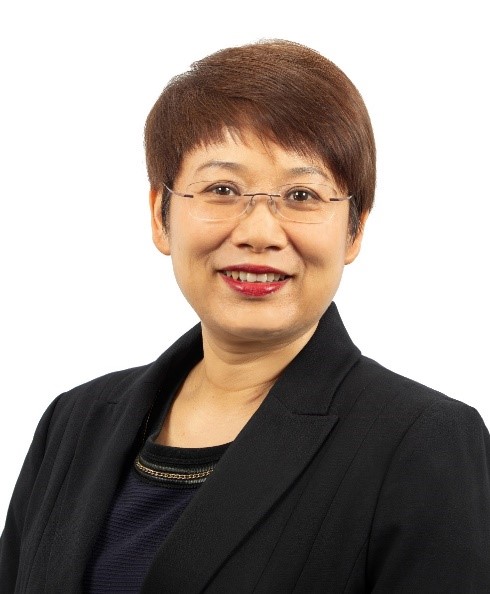}}]{Sumei Sun} is a Principal Scientist and Head of the Communications and Networks Dept at the Institute for Infocomm Research (I2R), Singapore. She is also holding a joint appointment with the Singapore Institute of Technology, and an adjunct appointment with the National University of Singapore, both as a full professor. Her current research interests are in next-generation wireless communications, cognitive communications and networks, and industrial internet of things. She is Editor-in-Chief of IEEE Open Journal of Vehicular Technology, member of the IEEE Transactions on Wireless Communications Steering Committee, and a Distinguished Speaker of the IEEE Vehicular Technology Society 2018-2021. She’s also Director of IEEE Communications Society Asia Pacific Board and a member at large with the IEEE Communications Society. 
\end{IEEEbiography}
\vskip 0pt plus -1fil 	
	
\end{document}